%% file: acl_latex.tex
\documentclass[11pt]{article}

\usepackage[preprint]{acl}

\usepackage{times}
\usepackage{latexsym}
\usepackage{enumitem}
\usepackage[T1]{fontenc}

\usepackage[utf8]{inputenc}

\usepackage[nopatch=footnote]{microtype}

\usepackage{inconsolata}

\usepackage{graphicx}

\usepackage[table, svgnames]{xcolor}

\usepackage{tabularx}
\usepackage{booktabs}
\usepackage{multirow}
\usepackage{makecell}
\usepackage{pifont}

\newcommand{\rx}{\textcolor{red}{\ding{55}}} 
\newcommand{\gm}{\textcolor{green}{\ding{51}}} 

\newcommand{\ignore}[1]{}

\usepackage[most]{tcolorbox}
\usepackage{varwidth}

\usepackage{lipsum}
\usepackage{setspace}
\usepackage{algorithm}
\usepackage{algorithmic}
\usepackage{amsmath}
\usepackage{amsfonts}
\usepackage{amssymb}
\usepackage{xspace}

\usepackage{listings}
\definecolor{lstbg}{RGB}{248,248,248}
\definecolor{lstcomment}{RGB}{117,113,94}
\definecolor{lstkeyword}{RGB}{0,0,180}
\definecolor{lststring}{RGB}{163,21,21}

\lstdefinestyle{mintedlike}{
  backgroundcolor=\color{lstbg},
  basicstyle=\ttfamily\small,
  keywordstyle=\color{lstkeyword}\bfseries,
  commentstyle=\color{lstcomment}\itshape,
  stringstyle=\color{lststring},
  numberstyle=\tiny\color{gray},
  stepnumber=1,
  numbersep=6pt,
  framerule=0.3pt,
  breaklines=true,
  breakatwhitespace=true,
  showstringspaces=false,
  tabsize=2,
  captionpos=b
}

\newcommand{\gptf}{\texttt{GPT-5}\xspace}
\newcommand{\bench}{\texttt{VRRPI-Bench}\xspace}
\newcommand{\diag}{\texttt{VRRPI-Diag}\xspace}
\newcommand{\wu}{\texttt{WhatsUp}\xspace}

\definecolor{boxf}{HTML}{6096ba}

\newtcolorbox{insight}[1][Insight]{
  enhanced,
  breakable,
  colback=white,
  colframe=boxf,
  boxrule=0pt,
  borderline west={1.5pt}{0pt}{boxf},
  borderline east={0.6pt}{0pt}{boxf!60},
  borderline south={0.6pt}{0pt}{boxf!60},
  fonttitle=\bfseries\small,
  title=#1,
  left=6pt,
  right=4pt,
  top=3pt,
  bottom=4pt,
  before skip=6pt,
  after skip=6pt,
}

\title{Lost in Space? Vision-Language Models Struggle \\ with Relative Camera Pose Estimation}

\author{Ken Deng\textsuperscript{1, 2} \quad
    Yifu Qiu\textsuperscript{1} \quad 
    Yoni Kasten\textsuperscript{3} \quad 
    Shay B. Cohen\textsuperscript{1} \quad 
    Yftah Ziser\textsuperscript{3, 4} \\
    \textsuperscript{1}University of Edinburgh \quad 
    \textsuperscript{2}University of Oxford, \\
    \textsuperscript{3}NVIDIA Research \quad 
    \textsuperscript{4}University of Groningen \\
    \texttt{ken.deng@stcatz.ox.ac.uk}, \texttt{\{yifu.qiu,scohen\}@ed.ac.uk}, \\
    \texttt{\{ykasten,yziser\}@nvidia.com}, \texttt{y.ziser@rug.nl}}

\begin{document}
\maketitle

\begin{abstract}
We study whether vision-language models (VLMs) can solve relative camera pose estimation (RCPE) from image pairs, a direct test of multi-view spatial reasoning. We cast RCPE as a discrete verbal classification task and introduce \texttt{VRRPI-Bench}, built from real RGB-D frames with object-centric camera motion, and \texttt{VRRPI-Diag}, which isolates individual motion degrees of freedom. Humans (0.91) and specialized geometric pipelines such as LoFTR (0.99) solve the task reliably, yet the best VLM reaches only 0.66 and most others remain near random. Our analyses show that this gap is not basic spatial competence: strong VLMs are near ceiling on single-image benchmarks, but most remain near random once reasoning must span views. They are unstable under source-target reversal (best 59.7\% consistency) and remain weak even in simplified single-DoF settings, especially on optical-axis motions such as roll and depth translation (GPT-5: 0.46 on roll). These failures are useful: they localize concrete missing capabilities, namely cross-view correspondence, view-consistent reasoning, and projective camera-motion understanding, making RCPE a targeted diagnostic for improving multi-view spatial reasoning in VLMs.
\footnote{Code is available at \url{https://github.com/kdeng03/lost-in-space}}
\end{abstract}

\section{Introduction}

\begin{figure}[h!]
    \centering
    \resizebox{0.8\linewidth}{!}{
        \input{fig/bench_ex}
    }
    \caption{Examples from \bench. Camera moves from source observation (left) to target observation (right), with the verbalized camera motion for simultaneous translation and rotation around a shared object.}
    \label{fig:ex-bench}
\end{figure}

\begin{figure}[tb]
    \centering
    \input{fig/diag_ex}
    \caption{Examples from \diag. Camera moves from source viewpoint (left) to target viewpoint (right) and both positive and negative descriptions are provided.}
    \label{fig:ex-diag}
\end{figure}

Relative camera pose estimation (RCPE), which aims to recover the relative rotation and translation between two camera viewpoints, is a fundamental problem in computer vision and robotics~\cite{kelsey2006vision,zhou2008robot,elqursh2011line}. While a single image provides only a projective and inherently ambiguous view of the physical world, multiple images enable reliable 3D perception once their relative pose is known. Beyond its classical formulation, RCPE represents the minimal unit of motion estimation from visual observations, serving as a structural building block for higher-level perception systems, including navigation~\cite{cassinis2019review}, augmented reality~\cite{marchand2015pose}, and 3D reconstruction~\cite{mouragnon2006real,hoang2020object}, accurate estimation of relative camera motion is a prerequisite for system initialization, tracking, and re-localization.

Traditional approaches to RCPE rely on geometric constraints derived from low-level visual correspondences, such as keypoint matching and epipolar geometry~\cite{lowe2004distinctive,sun2021loftr}. While effective under moderate viewpoint changes, these methods often degrade under wide baselines, occlusions, or texture-poor regions. Humans, by contrast, remain relatively robust in these scenarios, suggesting that higher-level semantic and structural cues can help resolve spatial ambiguities. This observation motivates the question of whether models endowed with strong semantic priors can complement or extend classical geometry-based pipelines.

In parallel, VLMs have rapidly advanced and are increasingly deployed in embodied and interactive systems, particularly in mobile robotics~\cite{shah2023lm}.
Such systems operate over time and across changing viewpoints, requiring perceptual representations that remain consistent under camera motion. This naturally calls for reasoning about spatial relationships across multiple views rather than within a single static image. Recent studies have begun to evaluate spatial reasoning, but primarily focus on single-image 2D relations, synthetic scenes, or highly constrained 3D settings~\citep{kamath-etal-2023-whats,chen2024spatialvlm}. These benchmarks do not directly assess whether VLMs can reason about metric, multi-view geometry grounded in real images. As a result, modern VLMs' ability to infer camera motion remains unexplored.

We investigate whether state-of-the-art VLMs can infer relative 3D camera pose from pairs of real-world images. To facilitate this evaluation, we introduce \bench (\textbf{V}isual \textbf{R}easoning \textbf{R}elative \textbf{P}ose \textbf{I}dentification), a benchmark constructed from unlabeled egocentric videos with annotations describing verbalized changes in relative camera translation and rotation (\S\ref{sec:data-col}). The benchmark reflects realistic scenarios in which two views predominantly focus on the same object while undergoing simultaneous translation and rotation. To make RCPE compatible with current multimodal models, we formulate the task as a discrete classification problem that captures the dominant direction of camera motion rather than predicting a full continuous transformation. This formulation preserves the geometric essence of RCPE while enabling systematic evaluation of VLMs’ 3D and multi-view spatial reasoning abilities (See~\autoref{fig:taxonomy}).

Our experiments show existing VLMs (best $0.66$) lag behind both humans ($0.91$) and classical geometric methods ($0.99$) on RCPE (\S\ref{sec:res-bench}). We further demonstrate that VLMs are inconsistent (best $59.7\%$) when reasoning over image pairs, failing to preserve logical symmetry when source and target views are swapped (\S\ref{sec:consistency}). To understand the source of this gap, we introduce a diagnostic benchmark, \diag, which isolates single degree of freedom camera motions for fine-grained analysis (\S\ref{sec:alg}). Using this benchmark, we show that VLMs struggle most with transformations along the optical axis (\gptf achieves $0.90$ on average, but $0.46$ on roll prediction), particularly depth translation and roll rotation (\S\ref{sec:res-diag}).

Finally, we conduct an error analysis to disentangle perception, cross-view correspondence, and camera-perspective reasoning (\S\ref{sec:ea}). We show that while modern VLMs have largely solved basic single-image spatial relations (most VLMs over $90$, compared to $46.8$ by BLIP2), they fail to robustly track objects across viewpoints (most VLMs near random level $0.28$) and to internalize the inverse relationship between object motion and camera motion. Together, these findings indicate that current VLMs rely on shallow, image-plane heuristics rather than grounded 3D geometric reasoning. We argue that RCPE, and \bench in particular, provides a stringent and complementary testbed for evaluating spatial intelligence in VLMs. Progress on this task would signal meaningful advances in multi-view understanding and grounded 3D reasoning, which are prerequisites for deploying VLMs in real-world embodied systems.

\section{Related Work}
\label{sec:related-work}

\paragraph{Pose Estimation}
RCPE is a fundamental problem in computer vision, especially simultaneous localization and mapping (SLAM)~\cite{kelsey2006vision, zhou2008robot, elqursh2011line}. 
Classical methods are divided into two categories: 
(i) \textbf{Solver-based}: SIFT~\citep{lowe2004distinctive}, SURF~\citep{bay2006surf}, and LoFTR~\citep{sun2021loftr} are used to identify matching points across views, where a matrix solver~\citep{fischler1981random} computes the relative pose, emphasizing precision; 
(ii) \textbf{Learning-based}: \citet{rockwell20228, khatib2024leveragingimagematchingendtoend} use deep networks to estimate pose directly, focusing on robustness. FAR~\citep{Rockwell2024far} presents a hybrid approach, achieving state-of-the-art performance.

\paragraph{Enhancing Spatial Reasoning of VLM} The common trends in enhancing spatial reasoning capabilities can be summarized through several approaches:
(i) \textbf{Multi-modal Input}: SpatialRGBT~\citep{cheng2024spatialrgpt} and SSR~\citep{liu2025ssr} incorporate additional modules to handle extra-modal data, such as depth images and point clouds, to enhance 3D space understanding;
(ii) \textbf{Augmented-VLM}: SpatialPIN~\citep{ma2024spatialpin} and MM-Spatial~\citep{daxberger2025mm} introduce external tools to aid in constructing 3D scenes, while Cola~\citep{chen2023large} augments VLMs with an agent that acts as a coordinator;
(iii) \textbf{Training}: \textsc{VILASR}~\citep{wu2025reinforcing} uses both supervised fine-tuning and reinforcement learning, Spatial-MLLM~\citep{wu2025spatial} modifies the vision-encoder architecture and releases their pre-trained VLM, and SpatialCoT~\citep{liu2025spatialcot} uses supervised learning to align VLMs with coordinate representations for improved reasoning.

\begin{figure}
    \centering
    \includegraphics[width=\linewidth]{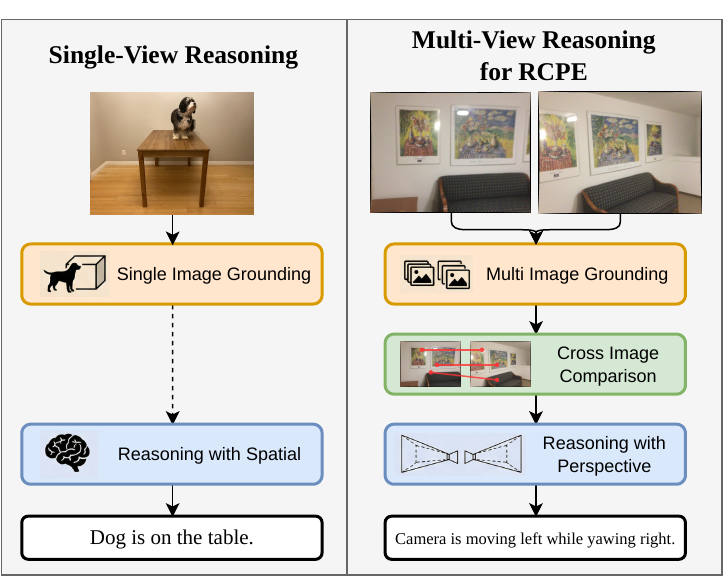}
    \caption{Capability taxonomy highlighting the skills required by \bench. Multi-view reasoning (right) raises new challenges in comparing with similar images, compared to single-view reasoning (left).}
    \label{fig:taxonomy}
\end{figure}

\paragraph{Spatial Reasoning Benchmark} Spatial reasoning in VLM is evaluated across several dimensions: 
(i) \textbf{2D Reasoning}: Sparkle~\citep{tang-etal-2025-sparkle} and SpatialEval~\citep{wang2024picture} build upon setups from LLM multi-hop reasoning tasks, treating image inputs as illustrations of grids, maps, or mazes; 
(ii) \textbf{3D Reasoning}: \citet{chen2024spatialvlm, daxberger2025mm, song2025robospatial, cai2025spatialbot} focus on understanding 3D scenes, emphasizing coordinate alignment, metric estimation~\citep{liao-etal-2024-reasoning}, and spatial relationships~\citep{kamath-etal-2023-whats, du2024embspatial}; 
(iii) \textbf{Mental Modelling}: Mind the Gap~\citep{stogiannidis2503mind} and MindCube~\citep{yin2025spatial} introduce a challenging task involving mental inference (reasoning on unobservable processes based on limited views);
(iv) \textbf{Comprehensive}: \citet{fu2024blink, shiri2024empirical, yang2025thinking} propose benchmarks that integrate the above tasks.
As illustrated in~\autoref{fig:taxonomy}, unlike previous spatial reasoning tasks, our benchmark involves cross-image comparison, where models must reconcile spatial relations across viewpoints. \bench evaluates a richer set of spatial reasoning skills than prior benchmarks. See App.~\ref{app:data-stat} for more comparison.

\section{Data Collection}
\label{sec:data-col}

\subsection{Preliminaries}
\label{sec:pre-geo}

\paragraph{Relative Pose Vector} We use \textbf{6 DoF} (degrees of freedom) to describe the camera's movement in a 3D scene: both rotation and translation along the $x$, $y$, and $z$ axes. For rotation, we use pitch ($\theta$) around the $x$-axis, yaw ($\phi$) around the $y$-axis, and roll ($\psi$) around the $z$-axis. For translation, we represent movement along the $x$-, $y$-, and $z$-axes as $t_x$, $t_y$, and $t_z$, respectively. Therefore, we use $\boldsymbol{v} \triangleq (\theta, \phi, \psi, t_x, t_y, t_z)$ to represent the relative pose vector, which corresponds to the 6 DoF. The rotation is expressed in radians
\footnote{Although rotations are represented in radians internally, we report and discretize angular differences in degrees for evaluation.}, while the translation follows the scale of the dataset metric units.

\paragraph{Intrinsic and Extrinsic Parameters} The camera's \textit{intrinsic matrix} is denoted as $\mathbf{K}\in \mathbb{R}^{3\times 3}$ and contains the pinhole camera projection parameters. The \textit{extrinsic matrix}, representing the camera's pose, is denoted as $\mathbf{T}_{wc}$, which is a homogeneous transformation matrix mapping points from the \textit{\underline{c}amera coordinate system} $\{c\}$ to a universal \textit{\underline{w}orld coordinate system} $\{w\}$.
Specifically, a point $\mathbf{p}_c$ expressed in $\{c\}$ is transformed to $\{w\}$ by $\mathbf{p}_w = \mathbf{T}_{wc}\mathbf{p}_c$.
The extrinsic matrix $\mathbf{T}_{wc}$ can be expressed as
 $\mathbf{T}_{wc} = \begin{bmatrix} \mathbf{R} & \mathbf{t} \\ \mathbf{0}^\top & 1 \end{bmatrix} \in \textit{SE}(3)$, 
where $\mathbf{R} \in \textit{SO}(3)$ is the rotation matrix and $\mathbf{t}\in \mathbb{R}^{3}$ is the translation vector.

\subsection{Filtering Algorithm}
\label{sec:alg}

Next, we describe two filtering schemes that we use in our experiments: \bench and \diag, representing global camera movement and single DoF change respectively. More technical details are provided in App.~\ref{app:data-col}.

\paragraph{Curation of \bench} For $i$-th RGB-D frame $\mathcal{F}_i$, we are given an RGB image $I_i$, a depth map\footnote{In depth map, each pixel value represents the camera-to-scene distance in a camera-dependent metric unit.} $D_i$, the camera intrinsic matrix $\mathbf{K}_i$, the camera index $c_i$, and the camera-to-world pose $\mathbf{T}_{wc_i}$, i.e., $\mathcal{F}_i = (I_i, D_i, \mathbf{K}_i, \mathbf{T}_{wc_i})$, we denote the entire dataset as $\{\mathcal{F}_i\}_{i=1}^{n}$, where all frames share a universal world coordinate system $w$. The filtering algorithm selects image pairs observing the same central object from different viewpoints, ensuring geometric consistency. For each pair $(I_i, I_j)$, we compute the angular difference $\tau$ and the mean reprojection distance $\bar{d}$ and keep it only if  $\bar{d} < d^{+}$, where $d^{+}$ is a predefined maximal reprojection error threshold. We then define the \textbf{angular ranges $\tau$} around a predefined angle set: $15^\circ,30^\circ,45^\circ,60^\circ$ representing different task levels and classify the pair into one of the segments. Intuitively, by maintaining the same 3D central point in the middle of both images, we ensure meaningful object-centric camera rotation. Note that such a movement usually requires a two DoF change in the extrinsic camera matrix.  This process produces viewpoint-consistent samples for \bench.

\paragraph{Curation of \diag} For $i$-th frame $\mathcal{F}_i$, we consider only the RGB image $I_i$ and its associated camera-to-world pose $\mathbf{T}_{wc_i}$, i.e., $\mathcal{F}_i = (I_i, \mathbf{T}_{wc_i})$. we denote the dataset as $\{\mathcal{F}_i\}_{i=1}^{n}$. Following the filtering principle of \bench, this algorithm selects image pairs that emphasize \textit{single-DoF motion} (translation or rotation along one axis). For each pair $(I_i, I_j)$, we compute the relative transformation $\mathbf{T}_{c_ic_j}$. A pair is retained if exactly one of the six DoF motions exceeds a threshold $\delta^{+}$, while the remaining DoF motions stay below $\delta^{-}$ (predefined per-DoF thresholds). This process produces single-DoF-motion samples for \diag.

\subsection{Overview}

The datasets that we use to collect data points are:
{
\newenvironment{itemizesquish}[2]{\begin{list}{\labelitemi}{\setlength{\itemsep}{#1}\setlength{\labelwidth}{#2}\setlength{\leftmargin}{\labelwidth}\addtolength{\leftmargin}{\labelsep}}}{\end{list}}

\begin{itemizesquish}{-0.3em}{0.5em}
\item \textbf{7 Scenes}~\citep{Shotton2013cvpr} 
is an indoor RGB-D dataset (\textasciitilde 43k) from Microsoft Research with seven environments and precise camera poses.

\item \textbf{ScanNet}~\citep{dai2017scannet} 
is a large-scale indoor dataset containing 2.5M frames across 1500 scenes with RGB-D information.

\item \textbf{ScanNet++}\footnote{Depth maps and intrinsics are unavailable for ScanNet++, so we only use it for \diag.}~\citep{yeshwanthliu2023scannetpp} contains 460 indoor scenes captured by laser scanners and commodity-level RGB-D cameras.
\end{itemizesquish}
}

\begin{table}[tb]
    \centering
    \small
    \input{tab/stat_bench}
    \caption{Statistics for \bench.}
    \label{tab:stat-bench}
\end{table}

Statistics about \bench are shown in~\autoref{tab:stat-bench}, while the statistics of \diag can be found in App.~\ref{app:data-stat} (\autoref{tab:stat-diag}).
We sample a maximum 300 samples per level for each dataset, resulting \textasciitilde 190 samples on average per split. 

\section{Experimental Setup}
\label{sec:experiment}

\paragraph{Task-Specific Baselines} We use SIFT~\citep{lowe2004distinctive} and LoFTR~\citep{sun2021loftr} as baseline methods for feature matching, followed by RANSAC~\citep{fischler1981random} to estimate the essential matrix and compute the relative pose.
SIFT identifies distinctive keypoints (unique locations within an image based on pixel values) that can be matched across different images, while LoFTR uses a vision transformer to directly match corresponding points (locations that appear in multiple images) between images. More details about evaluation can be found in App.~\ref{app:baseline-eval}

\paragraph{Vision-Language Model} We evaluate a range of VLMs and group them into four categories: 
1) \textbf{Open-source}: Idefics3~\citep{laurençon2024buildingbetterunderstandingvisionlanguage}, LLaVA-NeXT~\citep{liu2024llavanext}, LLaVA-OneVision~\citep{li2024llavaonevisioneasyvisualtask}, Qwen2.5-VL~\citep{bai2025qwen25vltechnicalreport}, and Qwen3-VL; 
2) \textbf{Proprietary}: GPT-4o~\citep{openai2024gpt4ocard} and GPT-5; 
3) \textbf{Fine-tuned VLM}: SpaceQwen~\citep{chen2024spatialvlm}, based on Qwen2.5-VL and fine-tuned on SpatialVLM; and 
4) \textbf{Reasoning VLM}: GLM-4.1V-Thinking~\citep{vteam2025glm45vglm41vthinkingversatilemultimodal} and Qwen3-VL-thinking, designed for step-by-step reasoning. See App.~\ref{app:vlm-eval} for prompts we use and evaluation details.

\paragraph{Human Evaluation} We randomly sample 40 examples from \bench and ask human annotators (25 graduate students) to follow a prompt matching the one provided to the VLMs,
without any supplementary information or examples. Further details on the prompt and evaluation procedure are provided in App.~\ref{app:human-eval}.

\paragraph{Evaluation} We treat the evaluation as \textit{binary classification}, and report \textbf{macro F1-score} to assess correctness and robustness of VLM performance.

\section{Results}

\subsection{Main Results on \bench}
\label{sec:res-bench}

Across both datasets, LoFTR sets the current performance upper bound, surpassing the SIFT baseline (\autoref{tab:res-bench}). LoFTR nearly saturates the 7 Scenes benchmark ($0.99$) and maintains high robustness on ScanNet ($0.92$). However, SIFT remains a strong baseline, outperforming the best-performing VLMs on both benchmarks. While competitive on 7 Scenes ($0.89$ average), its performance gap relative to LoFTR widens on the more challenging ScanNet dataset, where it scores $0.64$. Geometric methods show a pattern: performance drops with increasing viewpoint angles as local feature correspondences become harder. However, human annotators actually perform better at larger angles, as wider baselines require more global reasoning. VLMs, despite their semantic strengths, do not leverage global structural cues effectively to handle viewpoint shifts, often falling below random-level performance. This suggests that while humans use global priors to resolve spatial ambiguity, VLMs lack the geometric reasoning needed to exploit scene-level information. As a matter of fact, in several cases, where performance is low, the model returns degenerate answers, with a strong prior towards one answer that repeats and is also unaffected by the order of the images (\S\ref{sec:consistency}.

Model scale provides only modest improvements, with even the strongest open models, such as the largest Qwen variants, falling well short of classical baselines. \gptf stands as the clear exception, delivering the best VLM performance overall ($0.66$ on 7 Scenes and $0.64$ on ScanNet) and demonstrating the most stable behavior as the angle increases. On ScanNet, its accuracy at large rotations ($0.62$ to $0.66$) approaches that of traditional matchers, suggesting that VLMs show a potential in wide-baseline scenarios where geometric methods struggle. Despite this progress, a gap to human performance remains. Overall, the results emphasize both the current limitations of VLMs and the emerging potential of next-generation models in  large-angle settings.

\begin{table}[tb]
    \centering
    \scriptsize
    \setlength{\tabcolsep}{2pt}
    \resizebox{1.0\linewidth}{!}{
        \input{tab/main_res}

    }
    \caption{Macro-F1 results on \bench (for short: Lm - Llama, Lv - Llava, Q - Qwen, I - Instruct, T - Thinking, SQ - SpaceQwen), comparing CV method (top), VLM (mid) and human (bottom). Note that, for human evaluation, we test on a mixed subsets from both datasets, hence we duplicate the results here. \colorbox{red!25}{Red} represents the best score, while \colorbox{blue!25}{blue} shows the best score of VLMs. See App.~\ref{app:case-study} for case study.}
    \label{tab:res-bench}
\end{table}

\subsection{Consistency Analysis}
\label{sec:consistency}

\begin{figure}[tb]
    \centering
    \includegraphics[width=\linewidth]{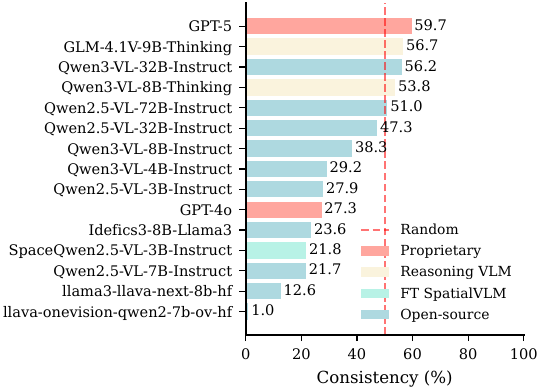}
    \caption{Consistency performance. The \textcolor{red!75}{dashed line} denotes the random baseline ($50.0\%$), highlighting that most of VLMs are around or below random level.}
    \label{fig:consistency}
\end{figure}

Previous research~\citep{zheng2023large,qiu-etal-2024-large, pezeshkpour2024large} has established that LLMs are sensitive to the position of information in a prompt. We follow this observation in the multimodal domain and check if VLMs are stable when the temporal or logical order of visual inputs is permuted.
We randomly sample 300 instances from \bench{} and swap the order of the two input images, treating the original source viewpoint as the target and vice versa, while keeping the textual prompt identical (the answer would be reversed). A model is considered consistent if it predicts the logically opposite option after the swap, regardless of the correctness of its initial prediction. As shown in \autoref{fig:consistency}, most VLMs fail to maintain this basic logical symmetry. Even \gptf, the top-performing model, achieves only 59.7\% consistency. This significant failure rate suggests that the internal visual representations in current VLMs are not robustly aligned with the language space, leading to a breakdown in the spatial-order information preserved between images.\footnote{We previously tested all VLMs with a sequence of a dog and a cat image, where they successfully recognized the image order. However, their performance dropped significantly when applied to 3D scene images and our high-level task.}

\begin{insight}
VLMs fail to internalize the logical symmetry of image pairs, indicating that their visual representations do not robustly encode spatial-order information relative to the language prompt.
\end{insight}

\subsection{Diagnostic Analysis by \diag}
\label{sec:res-diag}

The task in \bench involves composite spatial reasoning, requiring models to  infer both rotational and translational camera motions. To disentangle these components, we introduce \diag{}, evaluating a single DoF (a rotation or a translation), along one of the camera axes. 

\begin{figure}[tb]
    \centering
    \includegraphics[width=.85\linewidth]{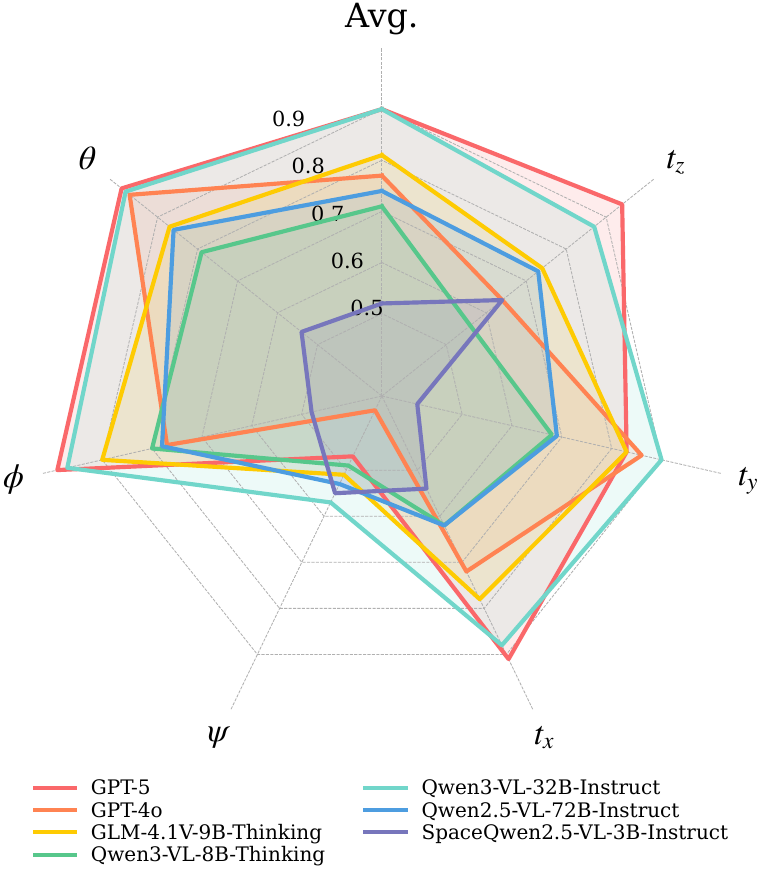}
    \caption{Results on \diag (top-performing VLMs of each category). The breakdown reveals a significant performance gap between image-plane shifts and optical-axis transformations. See full table in App.~\ref{app:tab-diag}.}
    \label{fig:res-diag}
\end{figure}

\autoref{fig:res-diag} shows models have gains when reasoning over isolated DoFs, yet performance is non-uniform across different axes. While VLMs are relatively proficient at identifying $x$ and $y$ translations, they struggle with transformations along the $z$-axis. We hypothesize that this discrepancy stems from a reliance on 2D motion heuristics; for small movements, transformations in the $xy$-plane often manifest as simple horizontal or vertical pixel shifts in the image plane, which models may track using low-level visual correspondences. 

In contrast, $z$-axis transformations, specifically depth (translation) and roll (rotation), require a deeper internalization of projective geometry. Translation along the $z$-axis induces ``looming'' effects and scale changes, while roll alters the canonical orientation of the  scene; neither can be approximated by simple 2D translations. This DoF-specific performance collapse suggests that VLMs rely on ``pseudo-spatial'' reasoning that fails when confronted with the unique projective distortions associated with the optical axis.

\begin{insight}
Testing isolated DoFs reveals a dual failure: VLMs suffer from a ``compositional gap'' when multiple motions are combined, yet their gains on easier datasets are largely driven by 2D image-plane heuristics that fail to generalize to the $z$-axis (depth and roll).
\end{insight}

\section{Error Analysis}
\label{sec:ea}

We use \diag to address three key questions: (i) Can VLMs recognize basic spatial relations within a single image (\S\ref{sec:ea-i})? (ii) Can VLMs compare two images to ground the \emph{same} object(s) and detect relational changes (\S\ref{sec:ea-ii})? (iii) Can VLMs improve in inferring the underlying camera-perspective motion when provided with textual descriptions (\S\ref{sec:ea-iii})?

\subsection{Intra-image Spatial Position Understanding}
\label{sec:ea-i}

\begin{table}[tb]
    \centering
    \scriptsize
    \setlength{\tabcolsep}{1pt}
    \resizebox{\linewidth}{!}{
        \input{tab/whatsup}
    }
    \caption{Performance on the WhatsUp dataset (left) and our annotated dataset (\textbf{bold col.} on the right). We compare current frontier VLMs against previous state-of-the-art models (e.g., BLIP-2) to establish a baseline for single-image spatial competence. See full table in App.~\ref{app:ea-i}.}
    \label{tab:whatsup}
\end{table}

To ensure that the failures with \bench{} are not merely due to lack of single-image spatial understanding, we revisit the findings of \citet{kamath-etal-2023-whats}, who showed that state-of-the-art models as of 2023 struggled to reason about even simple spatial relations within a single image. To verify that this limitation does not persist in current VLMs, explaining away their inability to perform more challenging tasks such as ours, we evaluate modern models on the \wu benchmark. This dataset focuses on 3D spatial relations (\textit{left} and \textit{right}) between objects within a single frame, where models are given image–caption pairs and must select the correct prepositional description from a candidate set. In addition, to ensure that our own data is not unsolvable due to similar low-level spatial ambiguities, we manually annotate a comparable single-image subset of \diag. Following \citet{kamath-etal-2023-whats}, we construct one-object tasks (determining an object’s position within the image: \textit{left, right, top, bottom}) and two-object tasks (determining the relative position between obj.1 and obj.2: \textit{left, right, above, below, front, behind}) over images drawn from 7 Scenes, ScanNet, and ScanNet++. Results on WhatsUp are shown on the left side of \autoref{tab:whatsup}, while results on our annotated \diag subset appear on the right.

Newer models do significantly better. While previous state-of-the-art models like BLIP-2 struggled with these fundamental relations, current frontier models such as \gptf and the larger variants of Qwen, demonstrate near-perfect performance, effectively reaching the ceiling of the WhatsUp benchmark. This suggests that the basic ability to resolve spatial prepositions within a static 2D context is largely a solved problem for modern VLMs. Consequently, the performance degradation observed in \bench cannot be attributed to a lack of basic spatial concepts, but rather to the inherent difficulty of 3D geometric reasoning across viewpoints.

\begin{insight}
    Modern VLMs have achieved near-saturation on single-image spatial benchmarks, significantly outperforming earlier models like BLIP-2. This confirms that their primary limitation lies not in basic spatial vocabulary, but in the higher-order geometric reasoning required for multi-view consistency.
\end{insight}

\subsection{Cross-image Relational Comparison}
\label{sec:ea-ii}

\begin{figure}[tb]
    \centering
    \includegraphics[width=\linewidth]{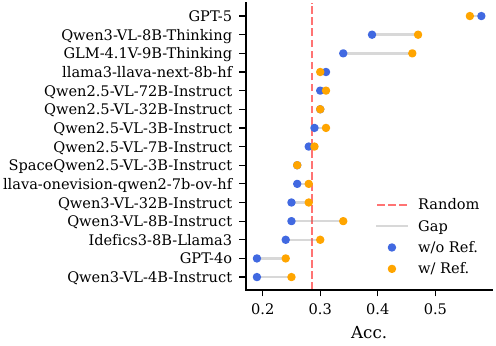}
    \caption{Cross-image relational comparison results. The \textcolor{red!75}{dashed line} denotes the random baseline ($\approx 0.28$). The \textcolor{gray!85}{bars} represent the ``Reference Gap'', the performance gain achieved when the target object is explicitly identified in the prompt. See App.~\ref{app:ea-ii} for full results.}
    \label{fig:ea-ii}
\end{figure}

Building on the intra-image baseline, we evaluate whether VLMs can maintain spatial consistency when comparing object relations across two disparate viewpoints. This task examines the model's ability to identify how the relative position of a shared reference object changes under camera motion. Given that 3D transformations can manifest as ambiguous visual shifts in 2D, we manually annotate each sample in \diag with all valid directional labels (e.g., \textit{left, right, upwards, closer}). A prediction is deemed correct if it matches any ground-truth label in the set.

To isolate the source of error, we use two prompting strategies:
(a) \textit{w/o Ref.}---the model must autonomously identify a reference object and infer its displacement; 
(b) \textit{w/ Ref.}---the reference object is explicitly provided, focusing the model's attention solely on the positional change.

\autoref{fig:ea-ii} reveals a systemic struggle with cross-view correspondence. While \gptf defines the current frontier with near-saturated performance, the majority of models, including \texttt{GPT-4o}, fail to significantly exceed the random baseline. A ``Thinking'' tier of models (e.g., Qwen3-VL-8B-Thinking) exhibits the largest performance gains when provided with an explicit reference. This indicates that for mid-tier models, the primary bottleneck is \textit{cross-view object correspondence}, the ability to recognize the same object across different projections. However, even with this assistance, most models remain far below human-level consistency, suggesting that the underlying geometric mapping remains fundamentally unresolved.

\begin{insight}
    While frontier models like \gptf begin to bridge the cross-view gap, most VLMs fail at the geometric mapping stage rather than simple object identification. The significant ``Reference Gap'' in reasoning-heavy models suggests that while they can resolve spatial changes, they lack the robust visual correspondence needed to autonomously link objects across 3D viewpoints.
\end{insight}

\subsection{Camera-perspective Reasoning}
\label{sec:ea-iii}

\begin{figure}[tb]
    \centering
    \includegraphics[width=\linewidth]{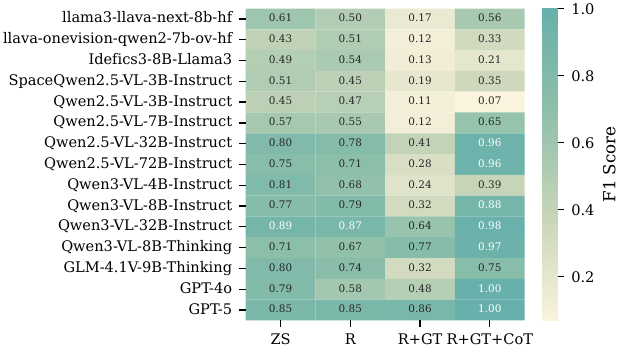}
    \caption{Camera-perspective reasoning. We evaluate whether models can synthesize object-level grounding into a global viewpoint transformation through four iterative setups: Zero-Shot (ZS), Ref Object Specification (R), Motion Grounding (R+GT), and Full Reasoning (R+GT+CoT).}
    \label{fig:ea-iii}
\end{figure}

Building on the previous diagnostic stages, we perform a sanity check to determine if the observed failures are rooted in a lack of perspective-transformation logic. We evaluate a subset of \diag under four iterative prompting setups: 
(1) \textbf{ZS}---the zero-shot baseline; 
(2) \textbf{R}---manually provided reference object identities; 
(3) \textbf{R+GT}---explicit object-motion annotations; and 
(4) \textbf{R+GT+CoT}---a complete reasoning chain describing the inverse relationship between object and camera motion.

In \autoref{fig:ea-iii}, several patterns emerge. First, providing object-level grounding (\textit{R}) fails to significantly improve performance, indicating that simply identifying relevant objects is insufficient for viewpoint reasoning. Second, the performance drop in \textit{R+GT} shows that despite being told exactly how an object moved, most VLMs fail to recognize that if a shared object appears to move in one direction across views, the camera must have moved in the opposite direction. This suggests a systemic inability to bridge the gap between egocentric object shifts and allocentric camera transformations.

Finally, the \textit{R+GT+CoT} setup acts as a critical control. While frontier models like \gptf reach near-perfect F1 scores when the full reasoning rule is explicitly stated in the prompt, this success appears to be a result of textual logic rather than visual understanding. By reducing the problem to a linguistic task, strong models can bypass the visual complexity. In contrast, smaller models are often distracted by the increased prompt complexity, leading to higher variance in their predictions.

\begin{insight}
The failure in the \textit{R+GT} setup confirms a specific reasoning deficit: VLMs struggle to internalize the inverse relationship between object-level visual shifts and global camera motion. While explicit linguistic rules (\textit{R+GT+CoT}) can compensate for this, it highlights a disconnect between the models' textual reasoning and their internal visual representations.
\end{insight}

We speculate that part of the challenge our benchmark poses to VLMs (versus the classic methods) is inherent in their architecture: they rely on coarse tokens to represent patches in the images, unaligned with the fine-grained nature of some of the tasks we present in our benchmark. This calls for future work in VLMs, where models become both sensitive to semantic, global information as well as low level (for example, at the pixel level) information.

\section{Conclusion}

We evaluated modern VLMs on RCPE and introduced \bench, a benchmark for realistic multi-view 3D spatial reasoning. Although humans and specialized geometric pipelines solve the task reliably, current VLMs remain far behind, and most are near random on the main benchmark. Our analyses show that this gap is not due to weak single-image spatial competence, but to failures in cross-view correspondence, view consistency, and projective camera-motion reasoning, especially for optical-axis transformations. We therefore view \bench not only as a challenging benchmark, but as a targeted diagnostic for the multi-view capabilities current VLMs still lack. Progress on this task would provide stronger evidence of grounded, view-consistent spatial reasoning in VLMs.

\section*{Limitations}


First, our benchmark focuses on indoor RGB-D environments, constructed from existing 3D datasets. While this design enables precise pose computation and controlled evaluation, the exploration in diverse environments including outdoor scenes or dynamic environments is limited.
Second, our study evaluates general-purpose VLMs in zero-shot setup. A potential direction is to investigate whether training objectives, such as geometric supervision or multi-view consistency constraints, can improve models’ ability to reason about camera motion.

\section*{Acknowledgements}

The authors acknowledge the use of resources provided by the Isambard-AI National AI Research Resource (AIRR). Isambard-AI is operated by the University of Bristol and is funded by the UK Government’s Department for Science, Innovation and Technology (DSIT) via UK Research and Innovation; and the Science and Technology Facilities Council [ST/AIRR/I-A-I/1023]~\citep{mcintoshsmith2024isambardaileadershipclasssupercomputer}. We also thank the Edinburgh Language Model team in their support through OpenAI credits.

\bibliography{acl_latex}

\appendix

\input{appendix}

\end{document}

%% file: fig/bench_ex.tex
\begingroup
\newtcolorbox{pairbox}[1][]{
  enhanced,
  colback=white,
  colframe=gray!50,
  boxrule=0pt,
  boxsep=0pt,
  left=0pt, right=0pt, top=0pt, bottom=0pt,
  sharp corners,
  title={#1}, 
  fonttitle=\bfseries,
  coltitle=black,
  colbacktitle=white,
  attach boxed title to top center,
  boxed title style={top=0pt, bottom=0pt, boxrule=.8pt},
}

\begin{tcbraster}[
    raster columns=1,
    raster equal height,
    raster column skip=0pt,
    raster row skip=0pt,
    enlarge top by=1pt,
    enlarge bottom by=1pt
]

\begin{pairbox}[$\tau \approx 15^\circ$ (Move left while yawing right \gm)]
    \centering
    \begin{minipage}[t]{0.48\linewidth}
      \centering
      \includegraphics[width=\linewidth]{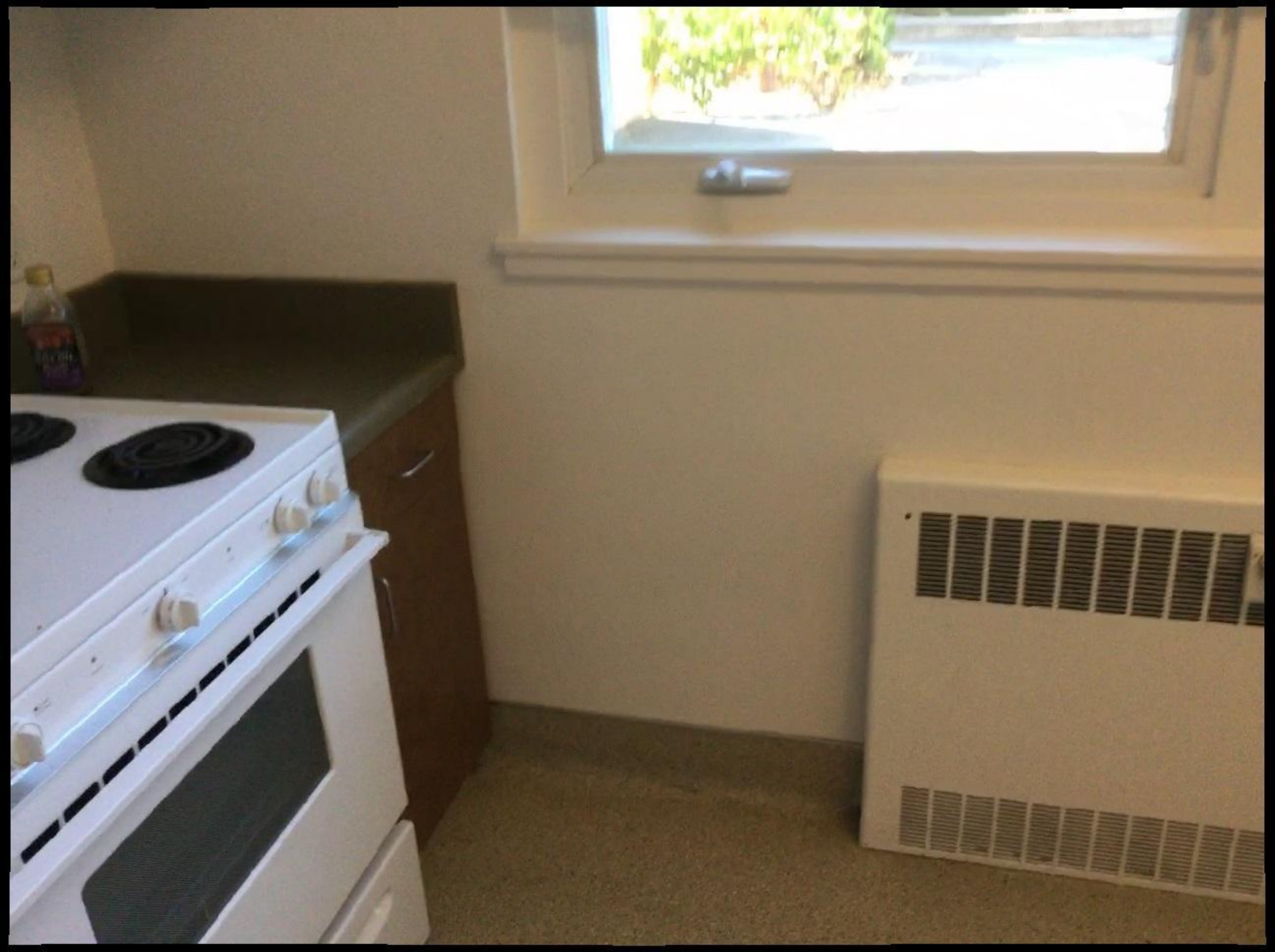}
    \end{minipage}
    \begin{minipage}[t]{0.48\linewidth}
      \centering
      \includegraphics[width=\linewidth]{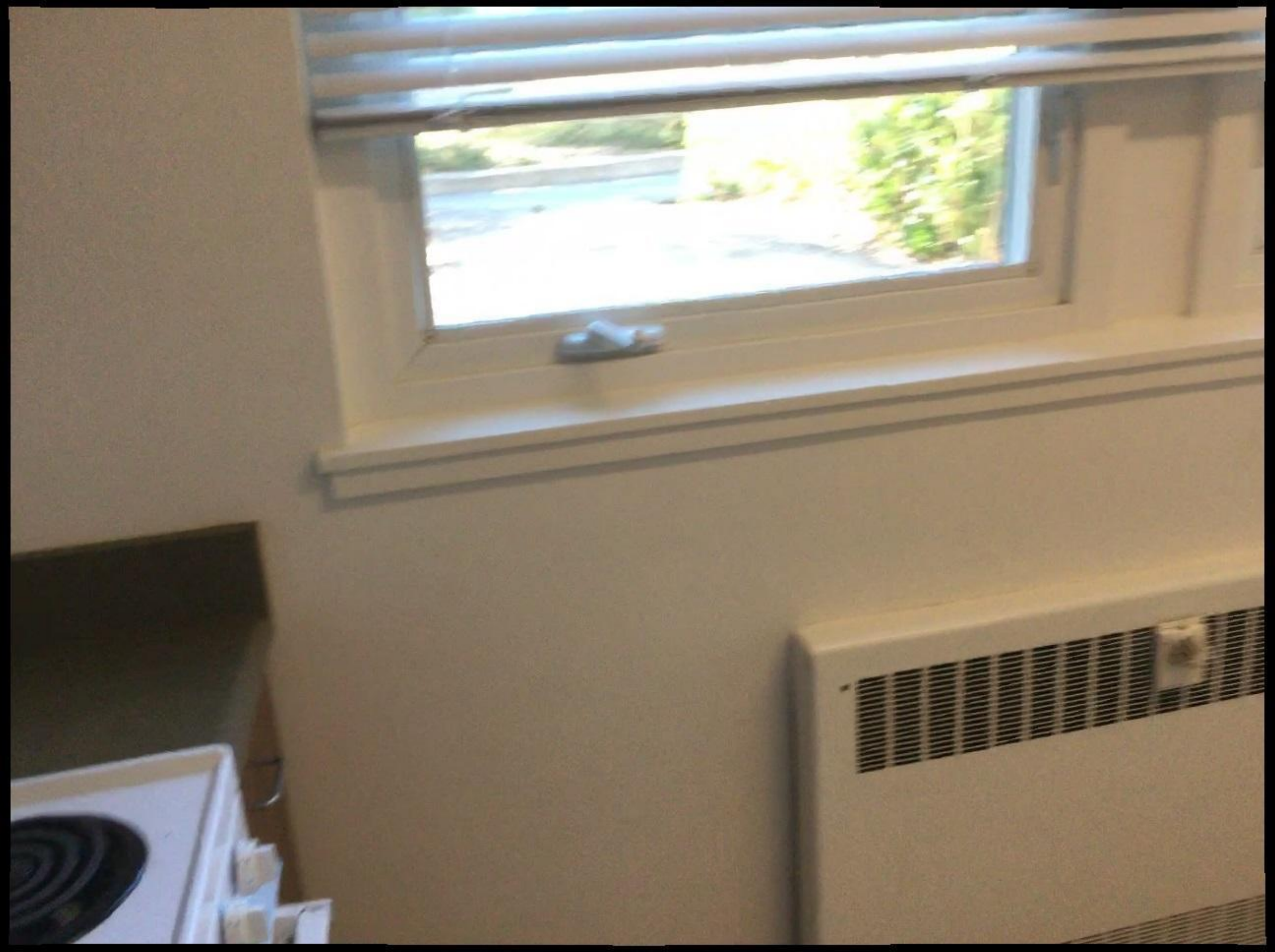}
    \end{minipage}
\end{pairbox}
\begin{pairbox}[$\tau \approx 30^\circ$ (Move left while yawing right \gm)]
    \centering
    \begin{minipage}[t]{0.48\linewidth}
      \centering
      \includegraphics[width=\linewidth]{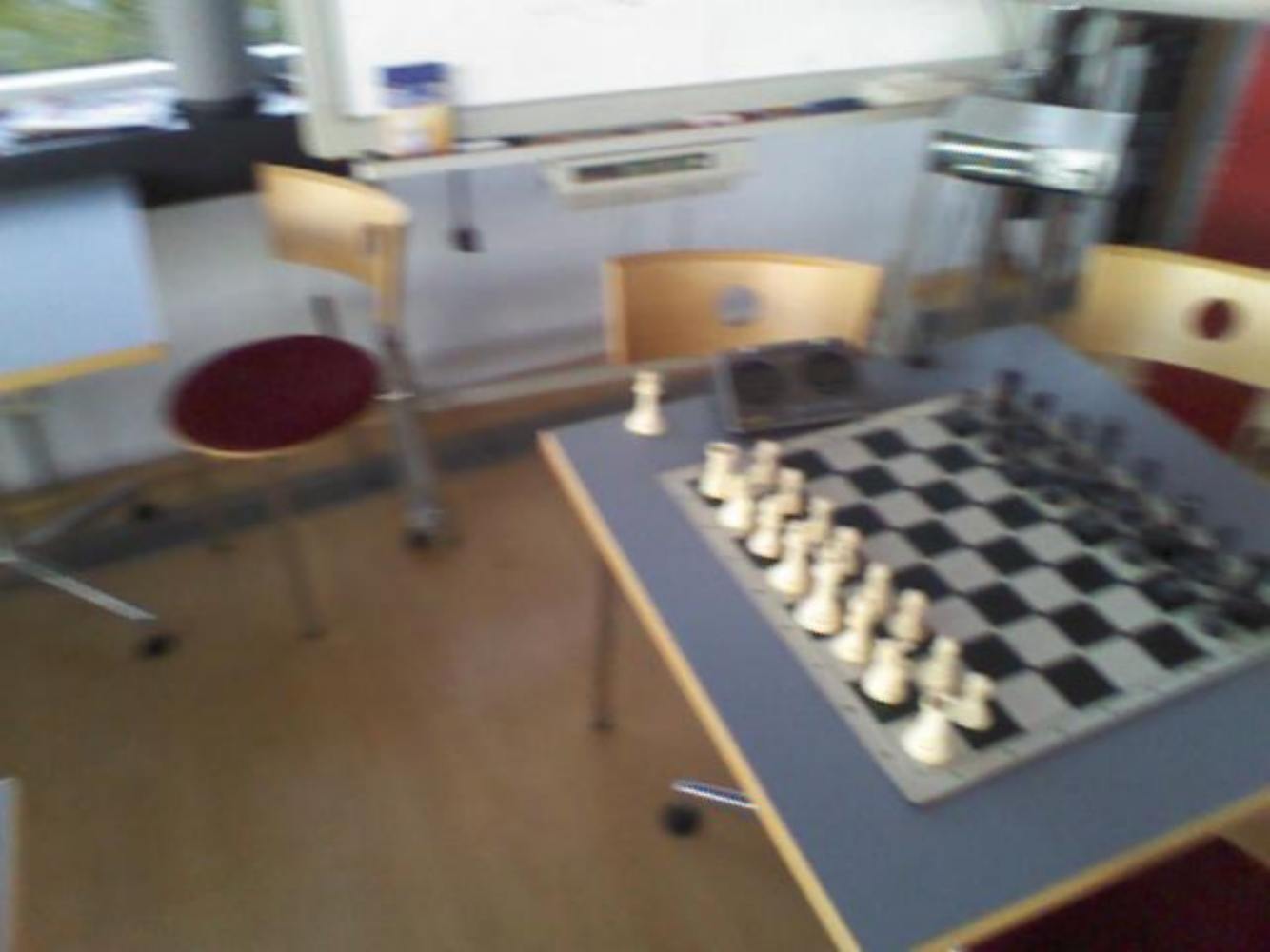}
    \end{minipage}
    \begin{minipage}[t]{0.48\linewidth}
      \centering
      \includegraphics[width=\linewidth]{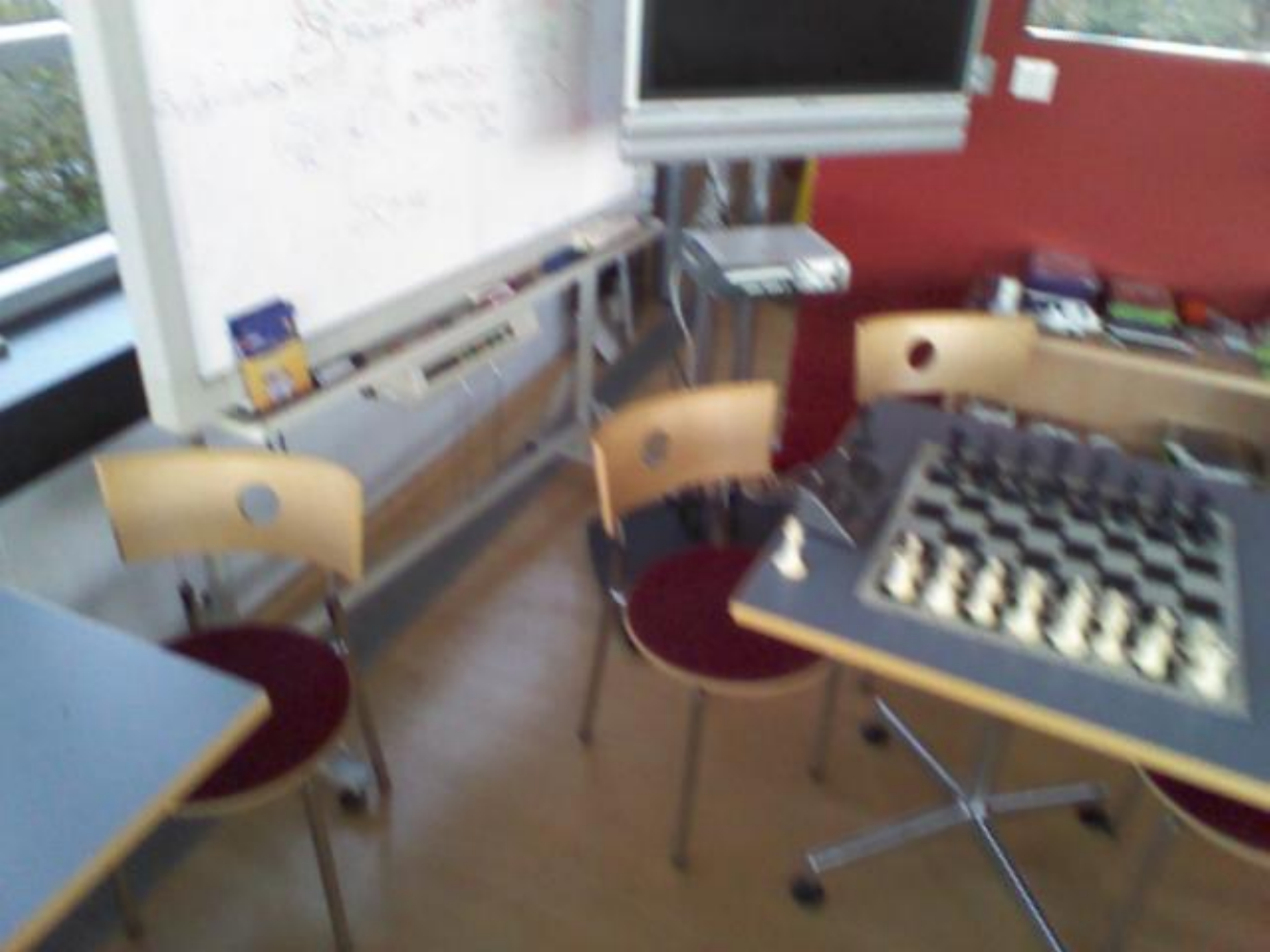}
    \end{minipage}
\end{pairbox}
\begin{pairbox}[$\tau \approx 45^\circ$ (Move left while yawing right \gm)]
    \centering
    \begin{minipage}[t]{0.48\linewidth}
      \centering
      \includegraphics[width=\linewidth]{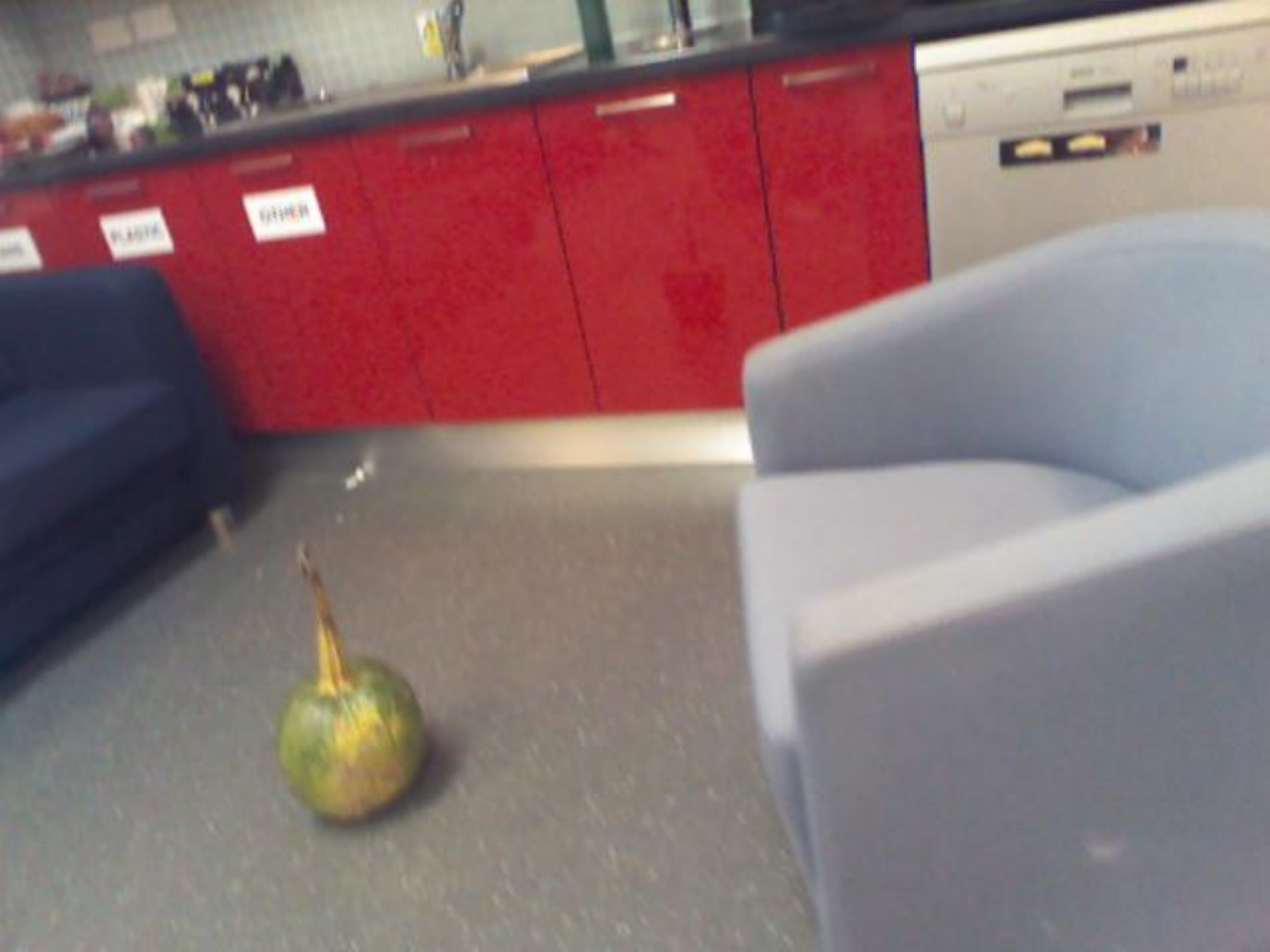}
    \end{minipage}
    \begin{minipage}[t]{0.48\linewidth}
      \centering
      \includegraphics[width=\linewidth]{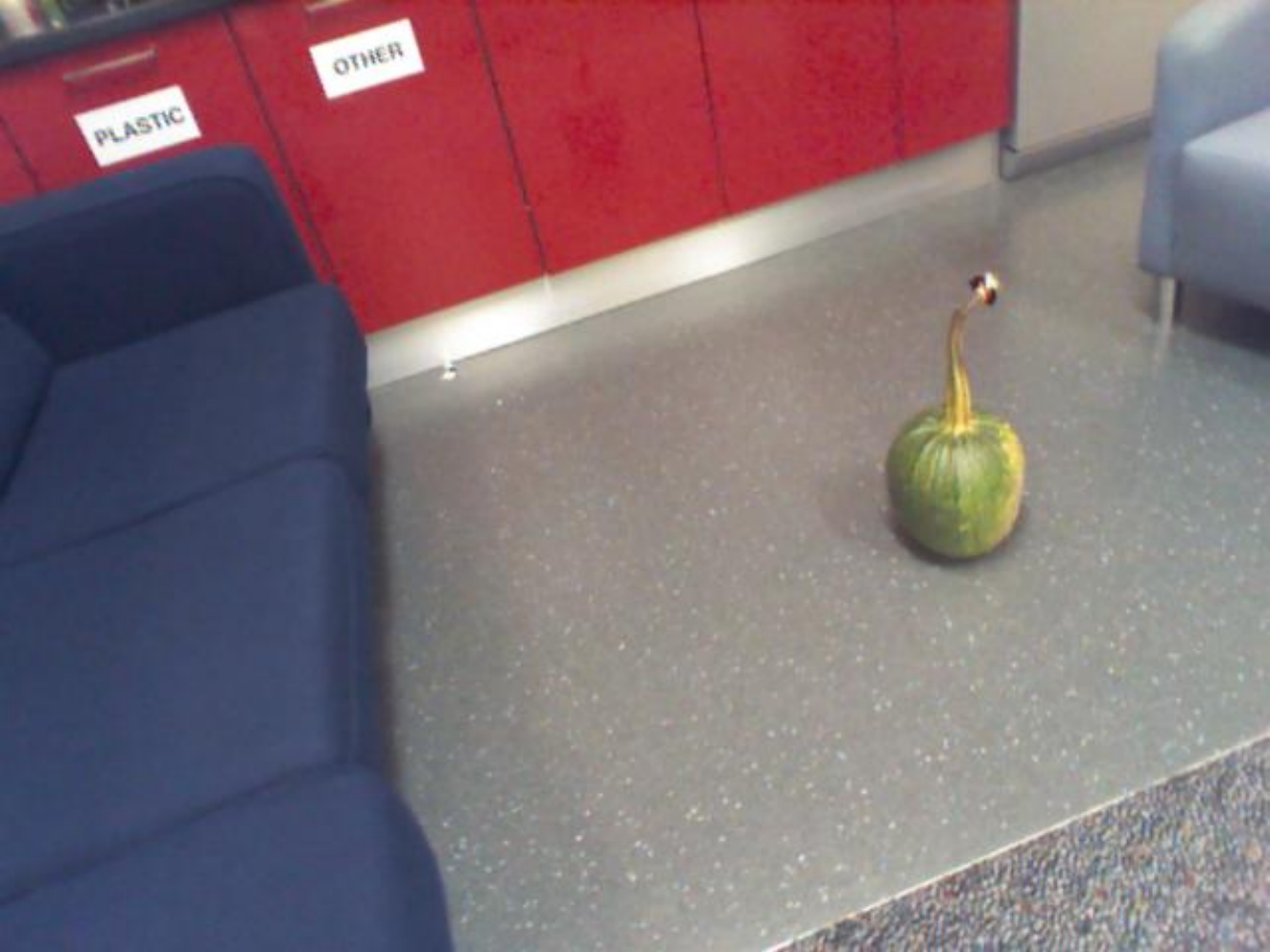}
    \end{minipage}
\end{pairbox}
\begin{pairbox}[$\tau \approx 60^\circ$ (Move right while yawing left \gm)]
    \centering
    \begin{minipage}[t]{0.48\linewidth}
      \centering
      \includegraphics[width=\linewidth]{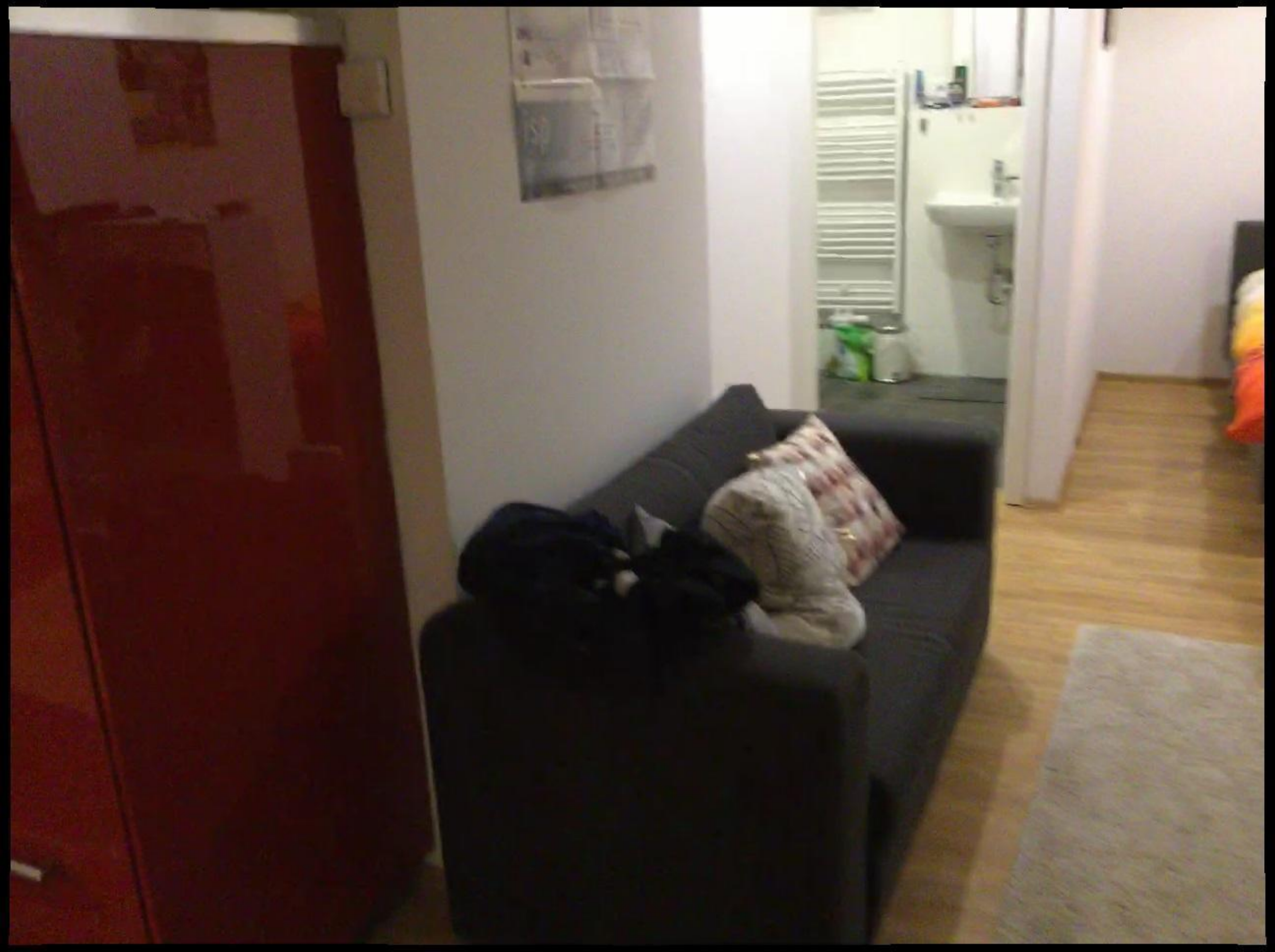}
    \end{minipage}
    \begin{minipage}[t]{0.48\linewidth}
      \centering
      \includegraphics[width=\linewidth]{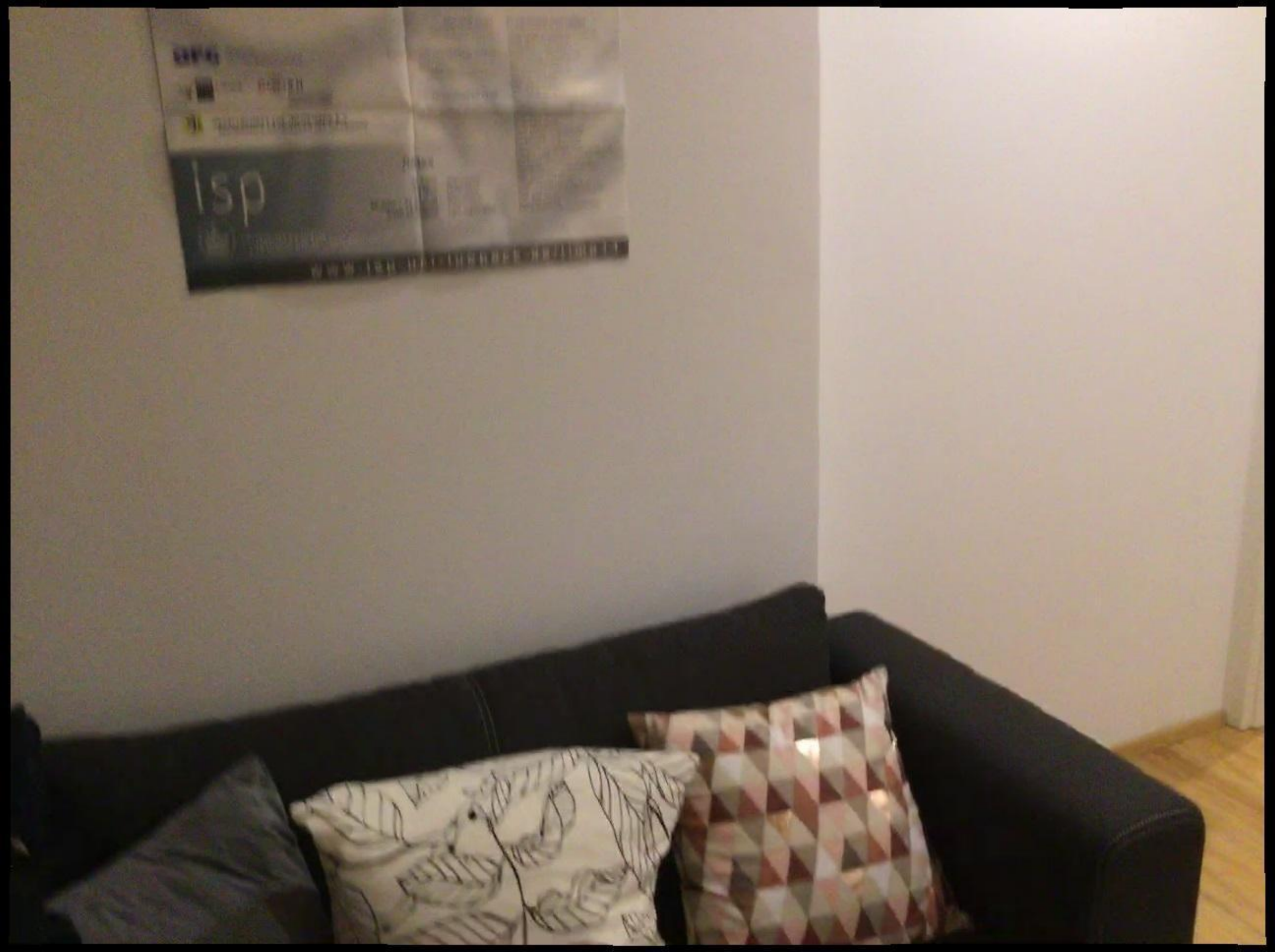}
    \end{minipage}
\end{pairbox}

\end{tcbraster}
\endgroup

%% file: fig/diag_ex.tex
\begingroup
\newtcolorbox{pairbox}[1][]{
  enhanced,
  colback=white,
  colframe=gray!50,
  boxrule=0pt,
  boxsep=0pt,
  left=0pt, right=0pt, top=0pt, bottom=0pt,
  sharp corners,
  title={#1}, 
  fonttitle=\bfseries\scriptsize,
  coltitle=black,
  colbacktitle=white,
  attach boxed title to top center,
  boxed title style={top=0pt, bottom=0pt},
}

\begin{tcbraster}[
    raster columns=2,
    raster equal height,
    raster column skip=0pt,
    raster row skip=0pt,
    enlarge top by=1pt,
    enlarge bottom by=1pt
]

\begin{pairbox}[$\theta:$ \parbox{13ex}{Pitch Up \gm \\ Pitch Down \rx}]
    \centering
    \begin{minipage}[t]{0.48\linewidth}
      \centering
      \includegraphics[width=\linewidth]{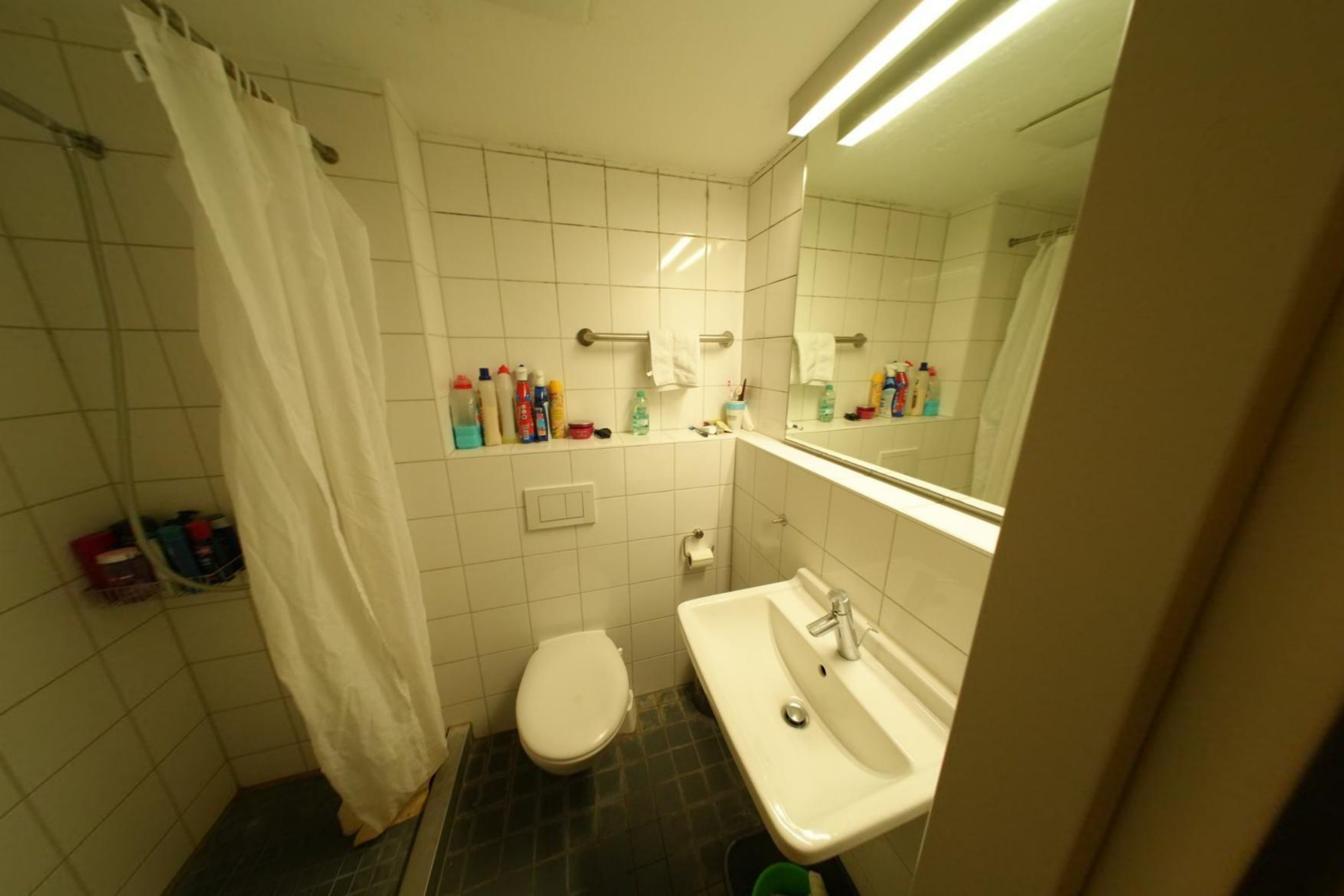}
    \end{minipage}
    \begin{minipage}[t]{0.48\linewidth}
      \centering
      \includegraphics[width=\linewidth]{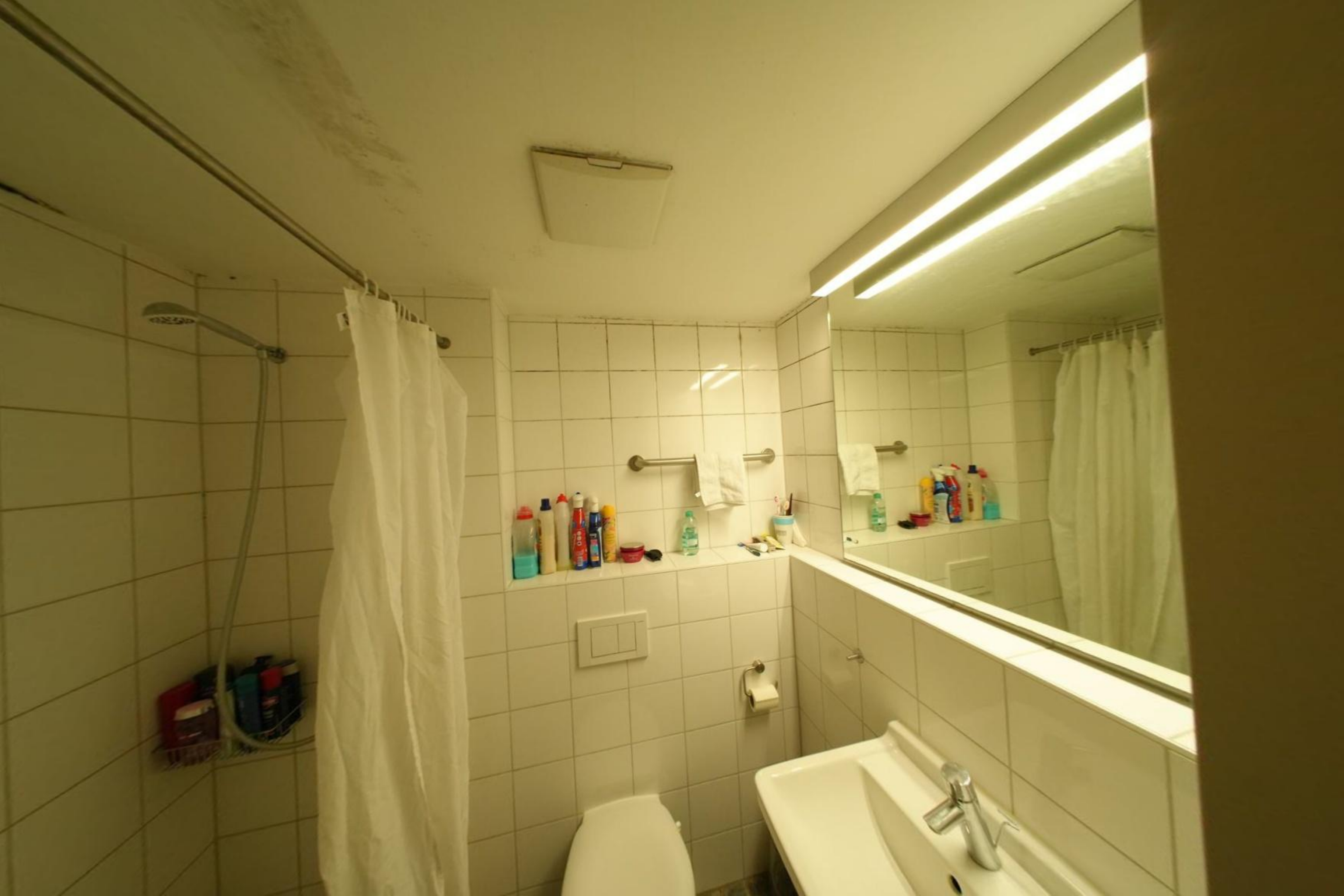}
    \end{minipage}
\end{pairbox}
\begin{pairbox}[$\phi:$ \parbox{11ex}{Yaw Left \gm \\ Yaw Right \rx}]
    \centering
    \begin{minipage}[t]{0.48\linewidth}
      \centering
      \includegraphics[width=\linewidth]{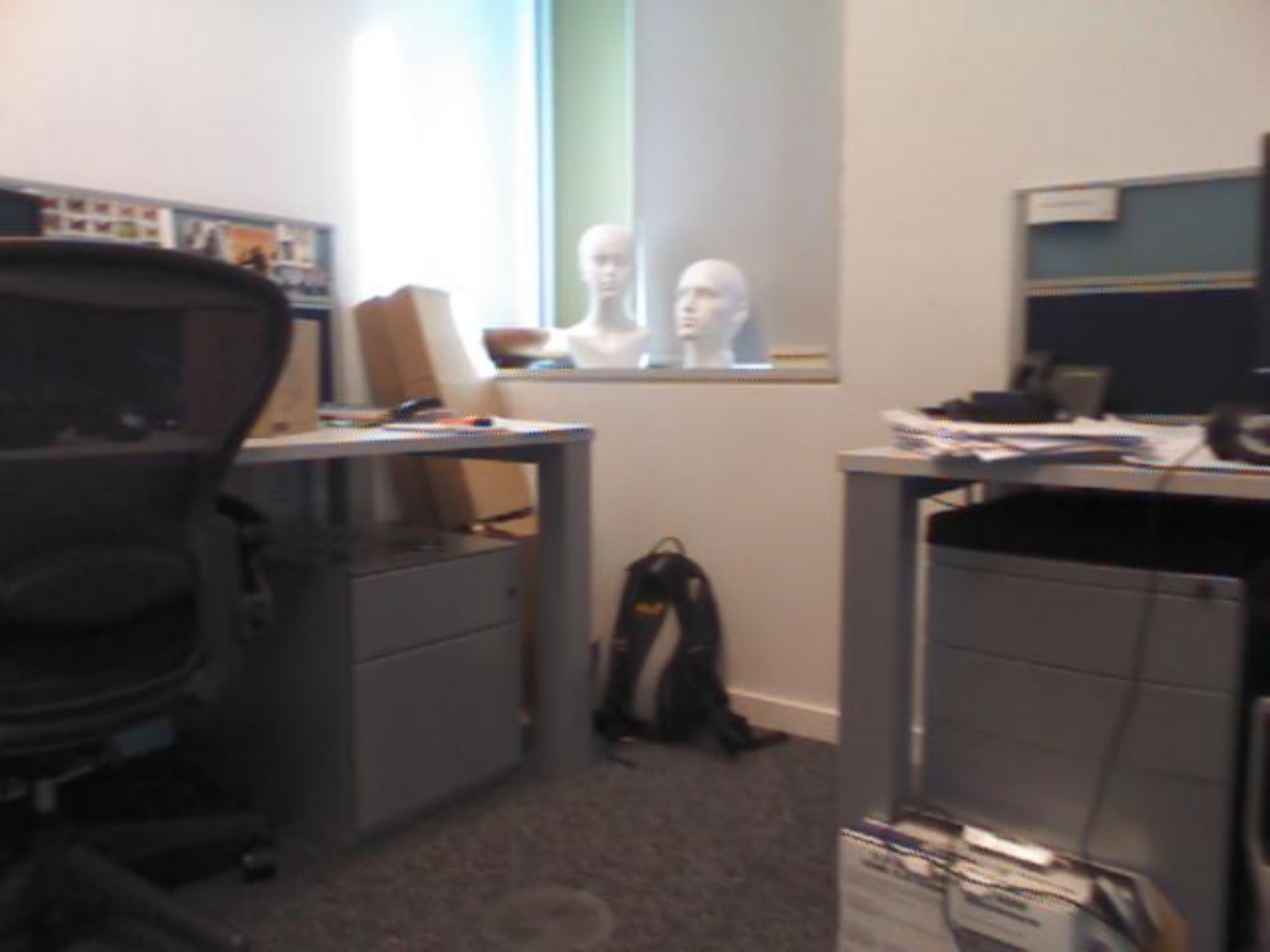}
    \end{minipage}
    \begin{minipage}[t]{0.48\linewidth}
      \centering
      \includegraphics[width=\linewidth]{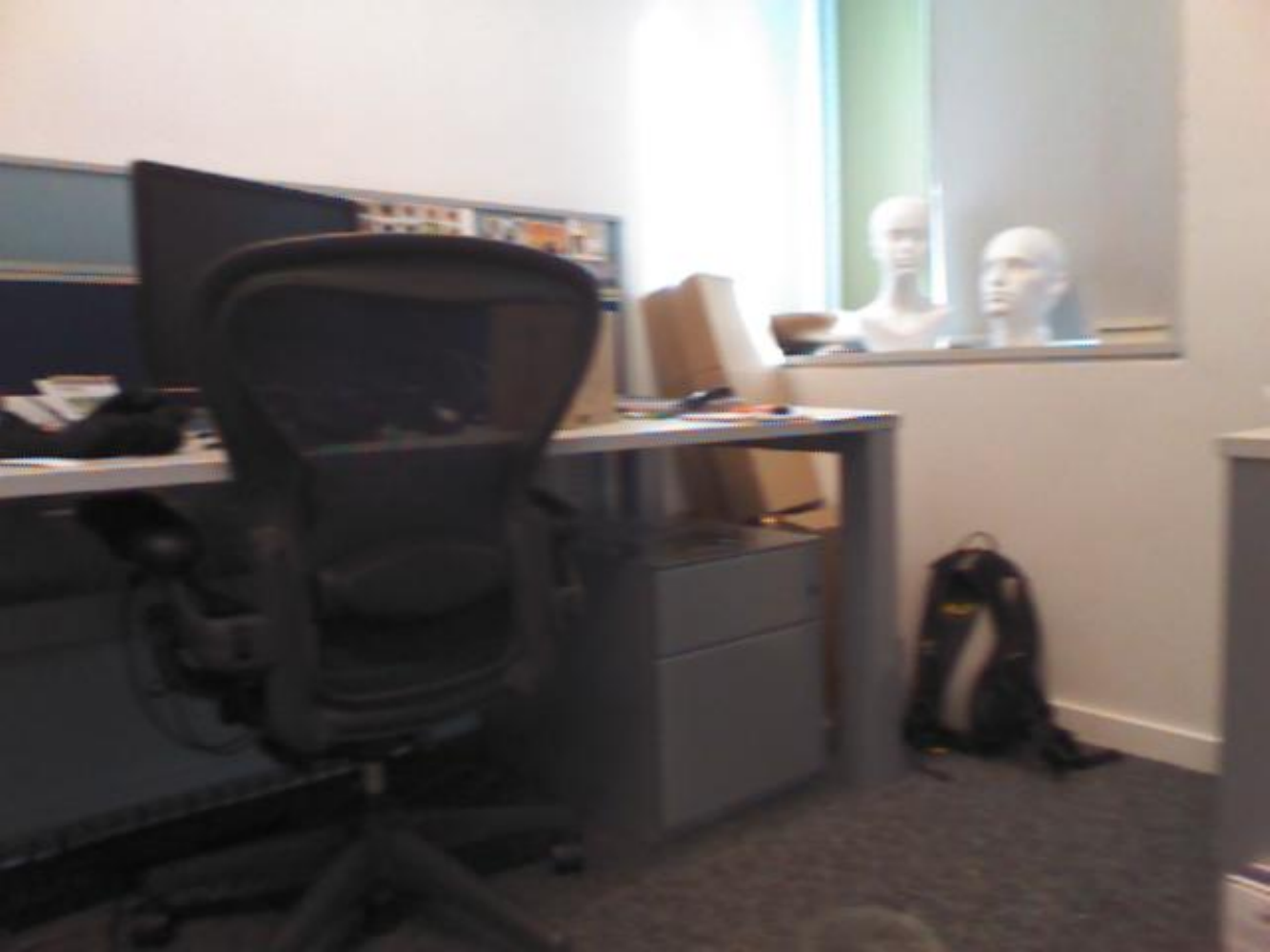}
    \end{minipage}
\end{pairbox}
\begin{pairbox}[$\psi:$ \parbox{15ex}{Roll C-Clock. \gm \\ Roll Clock. \rx}]
    \centering
    \begin{minipage}[t]{0.48\linewidth}
      \centering
      \includegraphics[width=\linewidth]{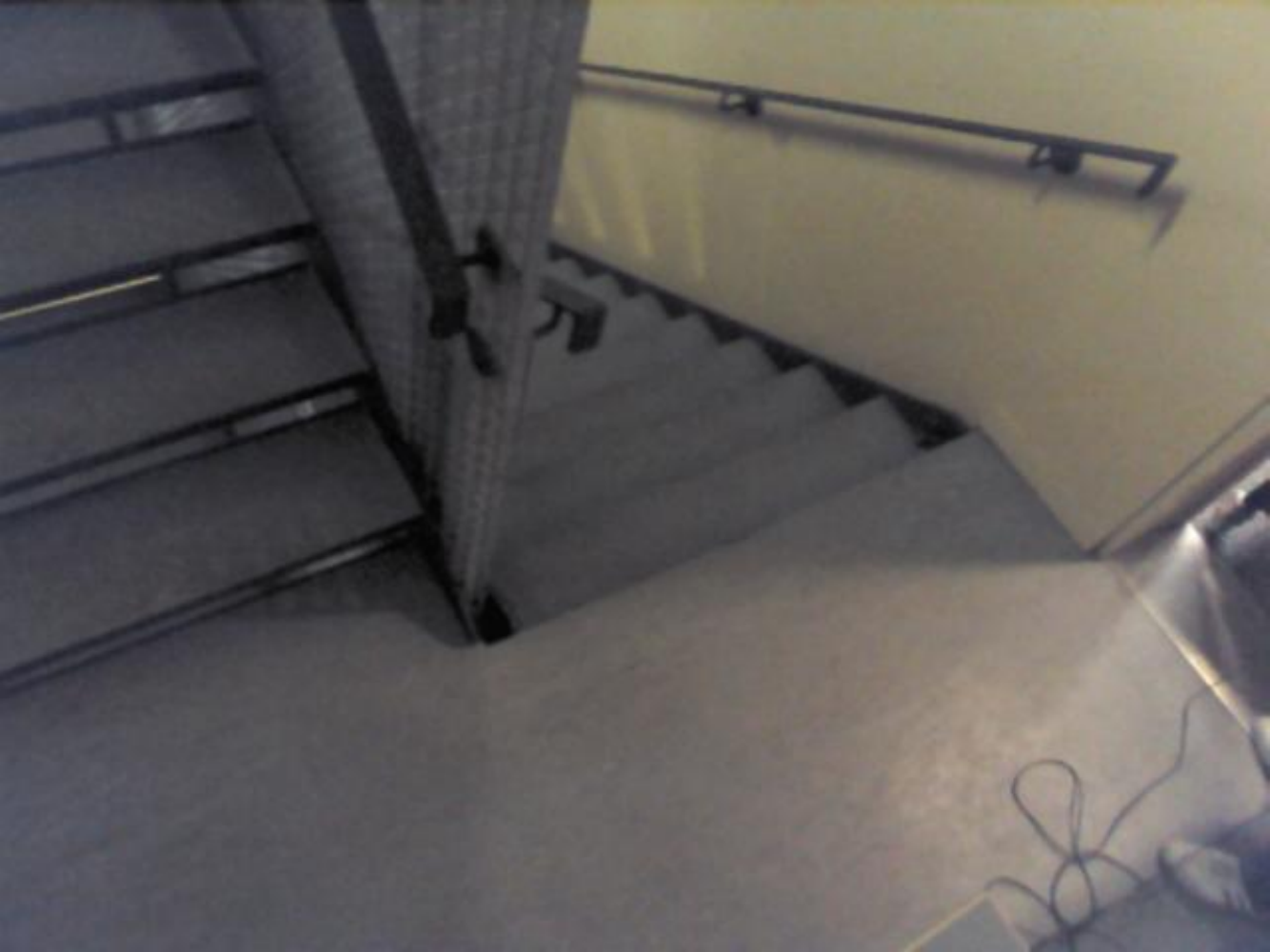}
    \end{minipage}
    \begin{minipage}[t]{0.48\linewidth}
      \centering
      \includegraphics[width=\linewidth]{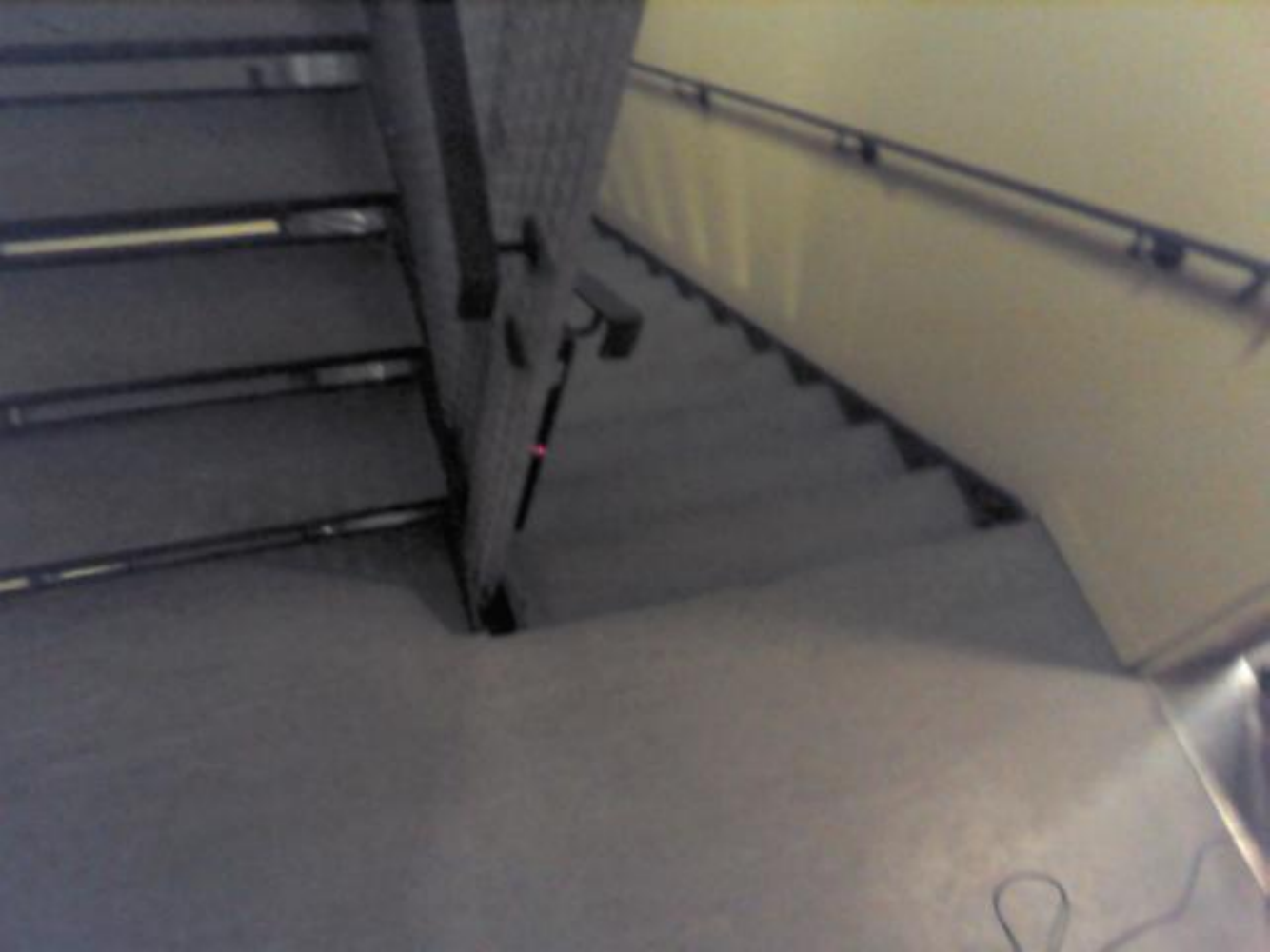}
    \end{minipage}
\end{pairbox}
\begin{pairbox}[$t_x:$ \parbox{14ex}{Trans. Right \gm \\ Trans. Left \rx}]
    \centering
    \begin{minipage}[t]{0.48\linewidth}
      \centering
      \includegraphics[width=\linewidth]{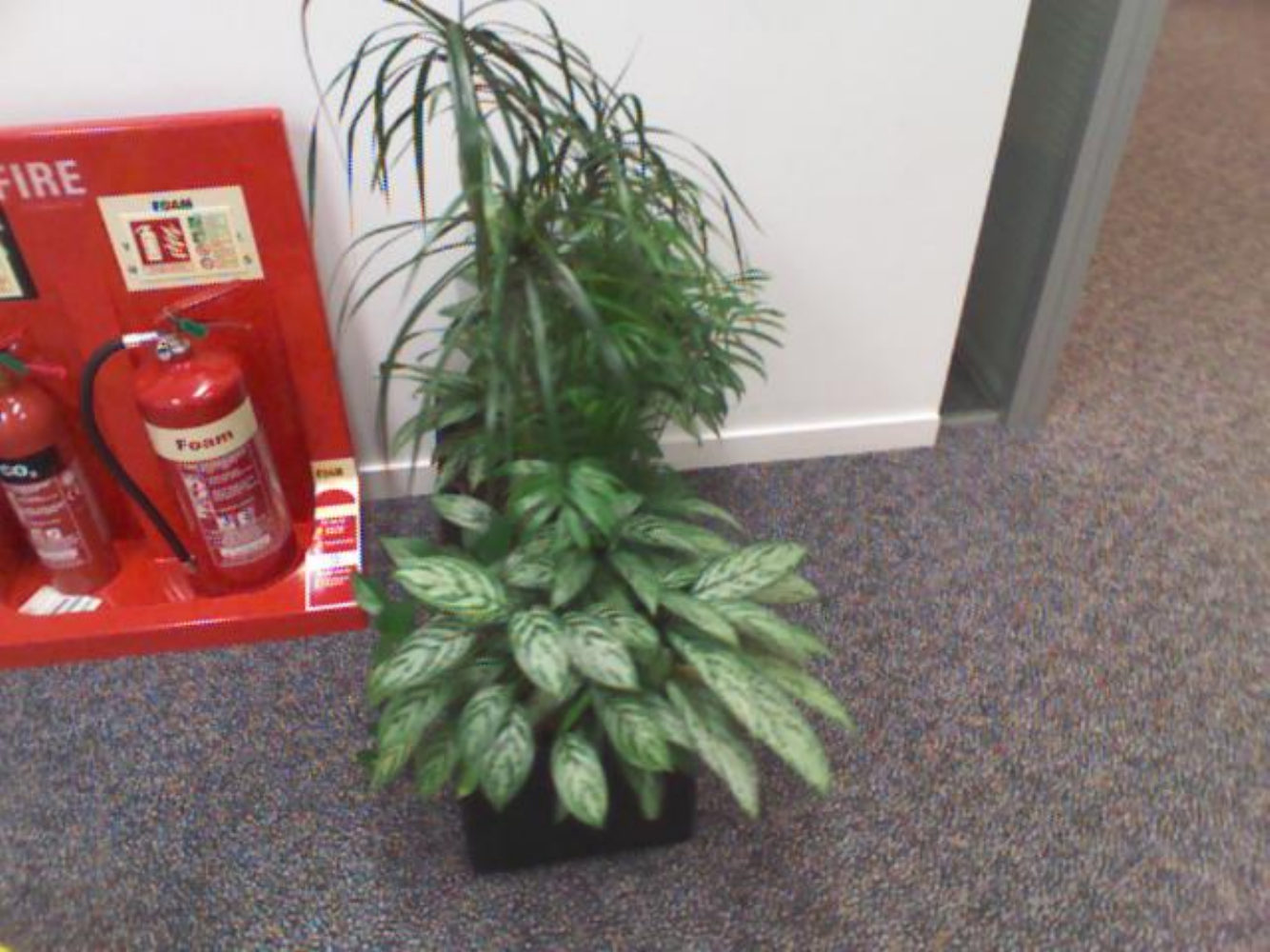}
    \end{minipage}
    \begin{minipage}[t]{0.48\linewidth}
      \centering
      \includegraphics[width=\linewidth]{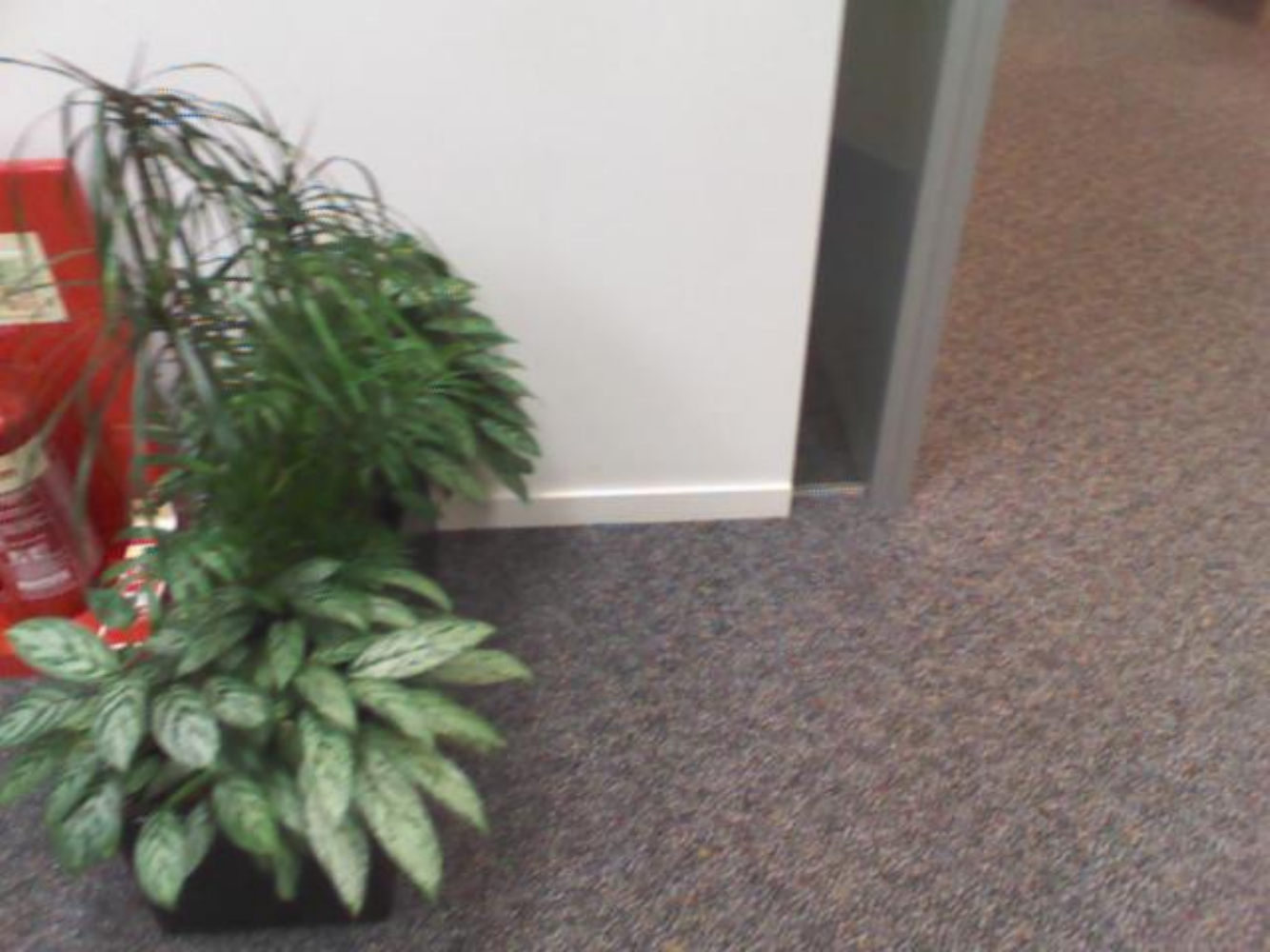}
    \end{minipage}
\end{pairbox}
\begin{pairbox}[$t_y:$ \parbox{14ex}{Trans. Down \gm \\ Trans. Up \rx}]
    \centering
    \begin{minipage}[t]{0.48\linewidth}
      \centering
      \includegraphics[width=\linewidth]{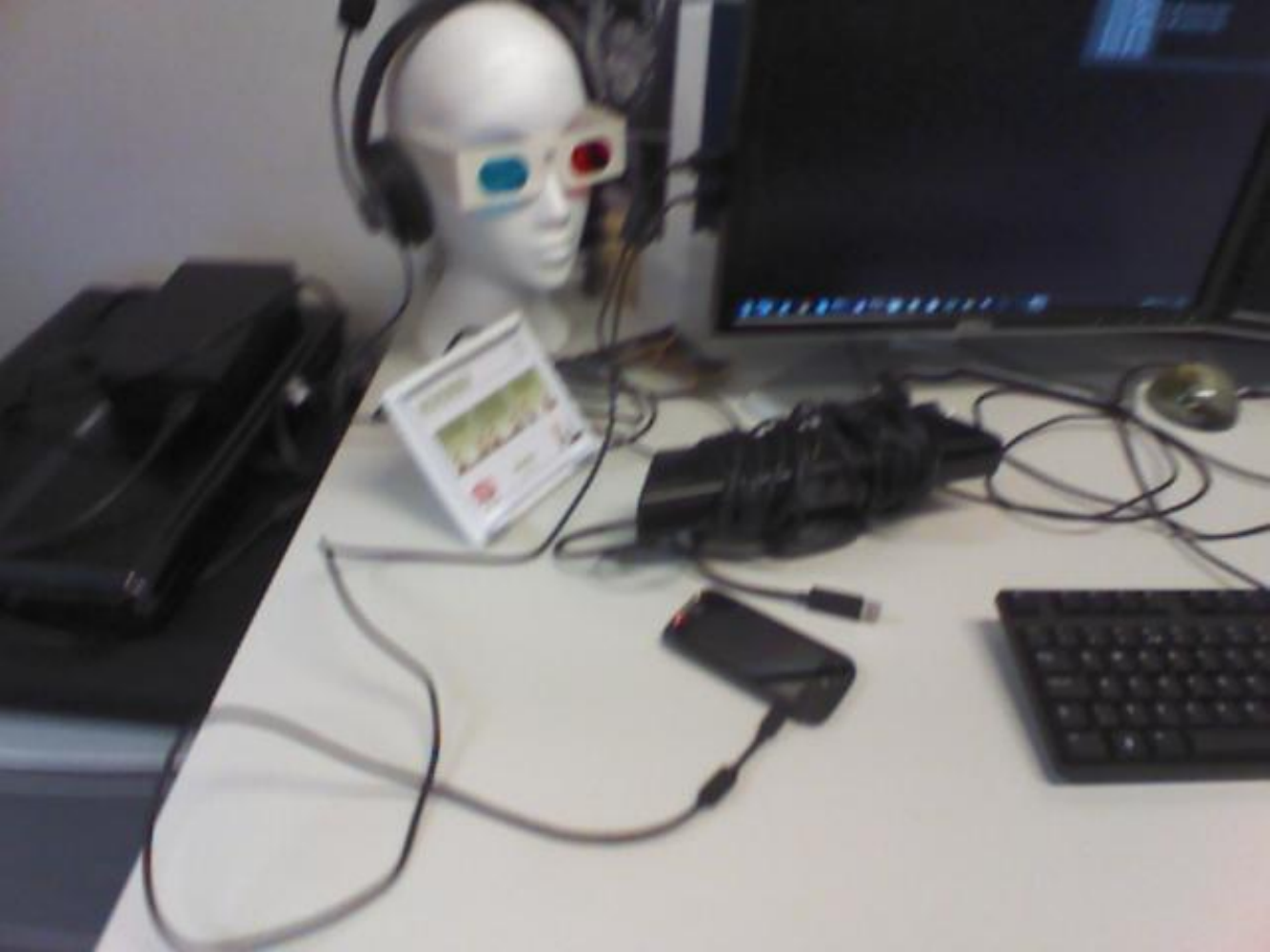}
    \end{minipage}
    \begin{minipage}[t]{0.48\linewidth}
      \centering
      \includegraphics[width=\linewidth]{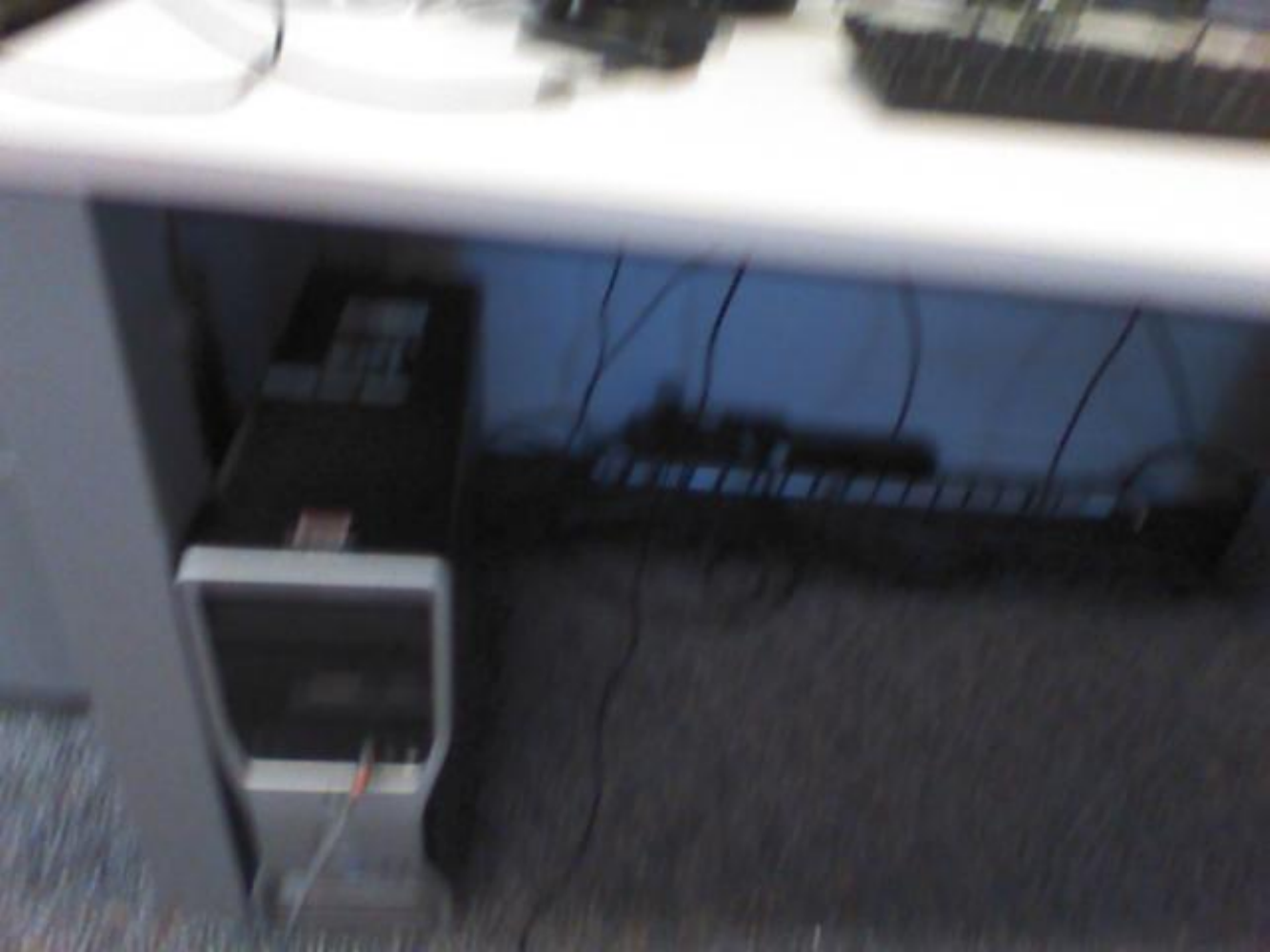}
    \end{minipage}
\end{pairbox}
\begin{pairbox}[$t_z:$ \parbox{18ex}{Trans. Forward \gm \\ Trans. Backward \rx}]
    \centering
    \begin{minipage}[t]{0.48\linewidth}
      \centering
      \includegraphics[width=\linewidth]{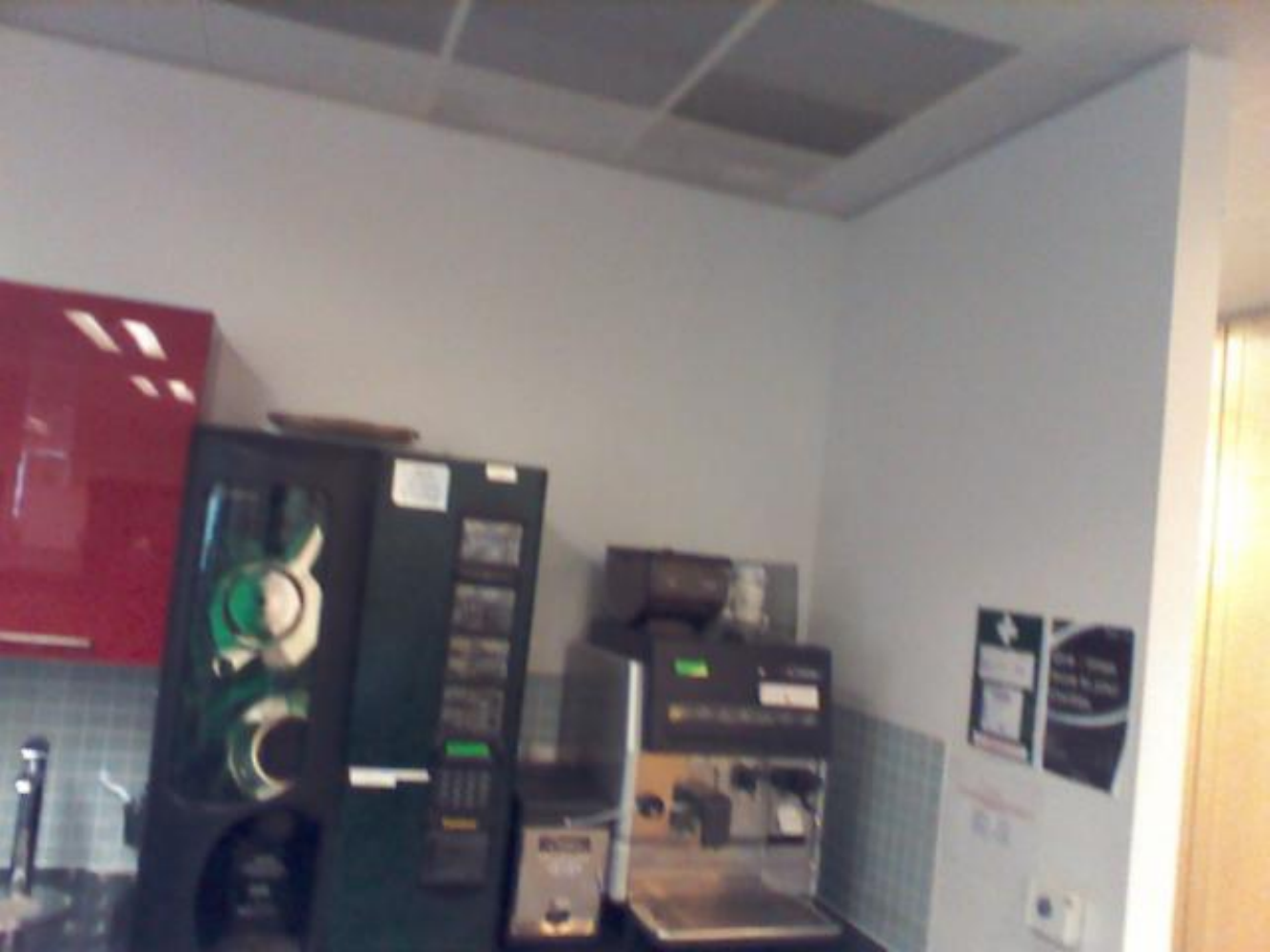}
    \end{minipage}
    \begin{minipage}[t]{0.48\linewidth}
      \centering
      \includegraphics[width=\linewidth]{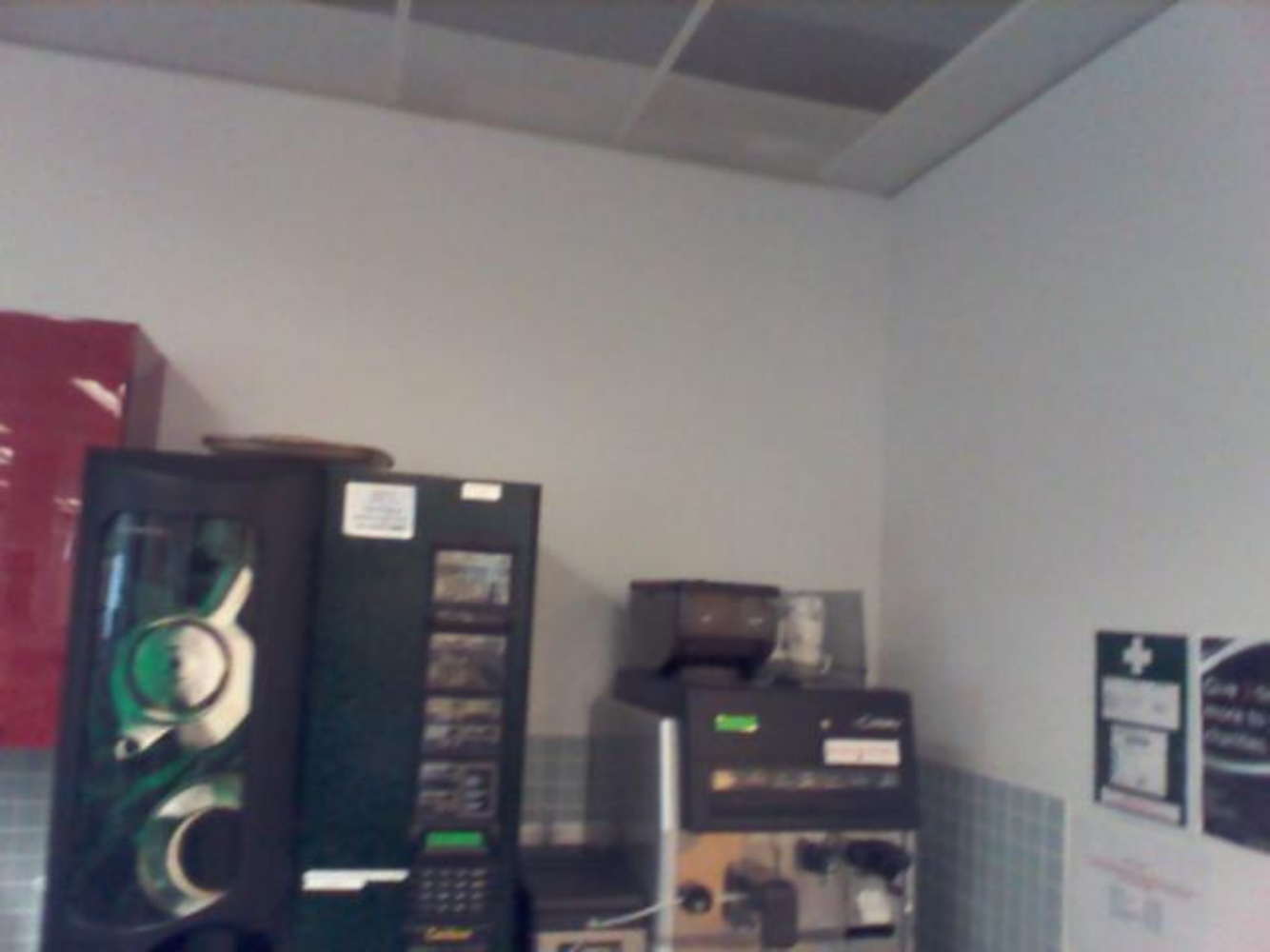}
    \end{minipage}
\end{pairbox}

\end{tcbraster}
\endgroup

%% file: tab/stat_bench.tex
\begin{tabular}{lcccc}
\toprule
 & \multicolumn{4}{c}{$\tau$} \\
 
 \cmidrule(lr){2-5}
 
 & \textasciitilde $15^\circ$ & \textasciitilde $30^\circ$ & \textasciitilde $45^\circ$ & \textasciitilde $60^\circ$ \\
 \midrule
\#~Frames & 564 & 413 & 350 & 186 \\
\bottomrule
\end{tabular}

%% file: tab/main_res.tex
\newcommand{\data}[2]{\multirow[c]{#1}{*}{\rotatebox{90}{\bfseries#2}}}

\newcommand{\best}[1]{\colorbox{red!25}{#1}}
\newcommand{\bestv}[1]{\colorbox{blue!25}{#1}}

\begin{tabular}{lccccc|ccccc}
\toprule
 & \multicolumn{5}{c|}{\textbf{7 Scenes}} & \multicolumn{5}{c}{\textbf{ScanNet}} \\
 \cmidrule(lr){2-6} \cmidrule(lr){7-11}
\textbf{Angle $\tau$} & $15^\circ$ & $30^\circ$ & $45^\circ$ & $60^\circ$ & Avg. & $15^\circ$ & $30^\circ$ & $45^\circ$ & $60^\circ$ & Avg. \\
\midrule
SIFT & 0.96 & 0.82 & 0.77 & 0.83 & 0.89 & 0.77 & 0.61 & 0.55 & 0.57 & 0.63 \\
LoFTR & \best{1.00} & \best{0.99} & \best{0.96} & \best{0.95} & \best{0.99} & \best{0.96} & \best{0.95} & 0.90 & 0.82 & \best{0.92} \\
\midrule
Lv-Next-8B-Lm3 & 0.40 & 0.43 & 0.42 & 0.38 & 0.41 & 0.42 & 0.40 & 0.40 & 0.38 & 0.40 \\
Lv-OneVision-7B-Q2 & 0.35 & 0.35 & 0.35 & 0.34 & 0.35 & 0.37 & 0.34 & 0.36 & 0.30 & 0.35 \\
Idefics3-8B-Lm3 & 0.47 & 0.48 & 0.43 & 0.40 & 0.47 & 0.49 & 0.51 & 0.52 & 0.53 & 0.51 \\
SQ2.5-VL-3B-I & 0.43 & 0.35 & 0.42 & 0.41 & 0.41 & 0.37 & 0.47 & 0.39 & 0.45 & 0.42 \\
Q2.5-VL-3B-I & 0.47 & 0.51 & 0.52 & 0.42 & 0.48 & 0.43 & 0.47 & 0.46 & 0.46 & 0.46 \\
Q2.5-VL-7B-I & 0.54 & 0.48 & 0.43 & 0.46 & 0.51 & 0.50 & 0.54 & 0.51 & 0.48 & 0.51 \\
Q2.5-VL-32B-I & 0.45 & 0.60 & 0.51 & 0.51 & 0.50 & 0.49 & 0.50 & 0.55 & 0.40 & 0.50 \\
Q2.5-VL-72B-I & 0.40 & 0.57 & 0.52 & 0.39 & 0.46 & 0.40 & 0.47 & 0.55 & 0.56 & 0.48  \\
Q3-VL-4B-I & 0.42 & 0.42 & 0.47 & 0.47 & 0.43 & 0.48 & 0.50 & 0.53 & 0.54 & 0.51 \\
Q3-VL-8B-I & 0.44 & 0.51 & 0.51 & 0.38 & 0.46 & 0.45 & 0.49 & 0.51 & 0.53 & 0.49 \\
Q3-VL-32B-I & 0.38 & 0.52 & 0.54 & 0.58 & 0.45 & 0.36 & 0.48 & 0.59 & 0.63 & 0.50 \\
Q3-VL-8B-T & 0.45 & 0.51 & 0.54 & 0.45 & 0.48 & 0.49 & 0.45 & 0.56 & 0.51 & 0.50 \\
GLM-4.1V-9B-T & 0.37 & 0.54 & 0.49 & 0.53 & 0.44 & 0.33 & 0.41 & 0.48 & 0.50 & 0.42 \\
GPT-4o & 0.40 & \bestv{0.63} & 0.50 & 0.56 & 0.48 & 0.38 & 0.43 & 0.50 & 0.53 & 0.45 \\
GPT-5 & \bestv{0.66} & \bestv{0.63} & \bestv{0.77} & \bestv{0.58} & \bestv{0.66} & \bestv{0.62} & \bestv{0.66} & \bestv{0.64} & \bestv{0.66} & \bestv{0.64} \\
\midrule
Human Annotators
& 0.88 & 0.84 & 0.92 & 0.92 & 0.91 & 0.88 & 0.84 & \best{0.92} & \best{0.92} & 0.91 \\
Random
& 0.50 & 0.50 & 0.50 & 0.50 & 0.50 & 0.50 & 0.50 & 0.50 & 0.50 & 0.50 \\
\bottomrule
\end{tabular}

%% file: tab/whatsup.tex
\begin{tabular}{lcccccc}
\toprule
 & \begin{tabular}{@{}c@{}}\texttt{Whats-} \\ \texttt{Up}\end{tabular} & \begin{tabular}{@{}c@{}}\texttt{COCO-} \\ \texttt{spatial}\end{tabular} & \begin{tabular}{@{}c@{}}\texttt{GQA-} \\ \texttt{spatial}\end{tabular} & Avg. & \begin{tabular}{@{}c@{}}\textbf{One-} \\ \textbf{obj}\end{tabular} & \begin{tabular}{@{}c@{}}\textbf{Two-} \\ \textbf{obj}\end{tabular} \\
\midrule
BLIP2-ITM & 37.6 & 53.0 & 49.8 & 46.8 & - & - \\
BLIP2-ITC & 29.0 & 53.7 & 51.0 & 44.6 & - & - \\
\midrule
Qwen2.5-VL-3B-Instruct & 81.5 & 90.0 & 90.8 & 87.4 & 74.5 & 42.7 \\
Qwen2.5-VL-32B-Instruct & 92.9 & 92.5 & 91.8 & 92.5 & 73.5 & 70.7 \\
Qwen3-VL-4B-Instruct & 98.7 & 85.5 & 88.9 & 91.0 & 80.3 & 66.3 \\
Qwen3-VL-32B-Instruct & 99.4 & 87.2 & 93.1 & 93.2 & 85.3 & 76.0 \\
GPT-4o & 72.8 & 63.3 & 61.2 & 65.7 & 64.8 & 55.8 \\
GPT-5 & 99.5 & 96.8 & 97.9 & 98.1 & 84.8 & 87.0 \\
\midrule
Random / Text-only & 25.0 & 50.0 & 50.0 & 41.7 & 25.0 & 16.7 \\
Human Estimate & 100.0 & 97.3 & 99.0 & 98.8 & - & -\\
\bottomrule
\end{tabular}

%% file: appendix.tex
\section*{Appendix}
\label{sec:app}

The list of appendices is as follows:
\begin{itemize}
    \item Data collection (App.~\ref{app:data-col}): Details about algorithm we use for creating datasets.
    \item Experiment details (App.~\ref{app:exp-setup}): Details about prompts and how we evaluate computer vision methods, VLMs and human annotators.
    \item Main results (App.~\ref{app:main-res}): Case study on \bench and full table result for \diag.
    \item Error analysis (App.~\ref{app:ea}): Full results on the WhatsUp dataset, together with visualization on our annotated dataset, and a full table for the cross-images grounding task.
\end{itemize}

\section{Data Collection}
\label{app:data-col}

\subsection{Geometry Computation}
\label{app:geo}

Below, we describe how to compute the relative pose vector $\boldsymbol{v}$ given two frames and their corresponding pose matrices, along with the angle $\tau$ and the central point deviation $d$ between the two viewpoints and their central intersection point.

\paragraph{Camera Coordinate System} We adopt the standard right-handed camera coordinate system, where the camera's optical center is located at the origin. The $z$-axis points forward, the $x$-axis points to the right, and the $y$-axis points downward.

\paragraph{Solve Relative Pose Vector from Extrinsic Parameters} Suppose extrinsic matrix for i-, and j-th camera system, $\mathbf{T}_{wc_i}$ and $\mathbf{T}_{wc_j}$, are given. If we want to determine how the $i$-th camera moves to the $j$-th camera, we aim to express the origin of the $j$-th camera's coordinate system in terms of the $i$-th camera's coordinate system. This can be achieved through a geometric transformation:
\begin{equation}
    \mathbf{T}_{c_ic_j} = (\mathbf{T}_{wc_i})^{-1} \mathbf{T}_{wc_j},
\end{equation}
where we can decompose $\mathbf{T}_{c_ic_j}$ into rotation matrix $\mathbf{R}$ and translation vector $\mathbf{t}$. The matrix $\mathbf{R}$ describes the rotation from the $i$-th camera to the $j$-th camera, and $\mathbf{t}$ represents the translation from the $i$-th camera to the $j$-th camera.

To obtain the relative pose vector $\boldsymbol{v}$, we use the following formulas from \citet{express_pose}:
\begin{align}
    \theta &= \arctan\left( \frac{\mathbf{R}_{32}}{\mathbf{R}_{33}} \right), \\
    \phi &= \arcsin\left( -\mathbf{R}_{31} \right), \\
    \psi &= \arctan\left( \frac{\mathbf{R}_{21}}{\mathbf{R}_{11}} \right).
\end{align}

Thus, $\boldsymbol{v}$ fully characterizes how the $i$-th camera moves relative to the $j$-th camera in absolute terms.

\paragraph{Unprojection and Reprojection} Unprojection refers to transforming a 2D coordinate into a 3D coordinate, while reprojection involves transforming a 3D coordinate into a 2D coordinate.

\paragraph{Angle ($\tau$) and Deviation Between Two Perspectives After Reprojection} Suppose we are given $\mathbf{T}_{wc}$ and $\mathbf{K}$ for the $i$-th and $j$-th cameras, along with the depth image $D$. Let $\boldsymbol{p}_i$ be the central point of the $i$-th image. To unproject it into the world coordinate, we use the following equation:
\begin{equation}
    \boldsymbol{p}_w = D_{\boldsymbol{p}_i} \mathbf{T}_{wc_i} (\mathbf{K}_i)^{-1} \boldsymbol{p}_i,
\end{equation}
where $D_{\boldsymbol{p}_i}$ denotes the depth value of the central point $\boldsymbol{p}_i$.

Next, we reproject this point onto the $j$-th image:
\begin{equation}
    \boldsymbol{p}_j = \mathbf{K}_j (\mathbf{T}_{wc_j})^{-1} \boldsymbol{p}_w.
\end{equation}

To compute the angle $\tau$ between the two viewpoints and the deviation $d$ of the central point after reprojection, we first decompose the origins of the $i$-th and $j$-th cameras, $\mathbf{o}_i$ and $\mathbf{o}_j$, in the world coordinate system using $\mathbf{T}_{wc_i}$ and $\mathbf{T}_{wc_j}$ . We then calculate:
\begin{align}
    \tau &= \text{arccos} \left( \frac{\vec{\boldsymbol{p}_w\mathbf{o}_i} \cdot \vec{\boldsymbol{p}_w\mathbf{o}_j}}{\|\vec{\boldsymbol{p}_w\mathbf{o}_i}\| \|\vec{\boldsymbol{p}_w\mathbf{o}_j}\|} \right), \\
    d &= \| \boldsymbol{p}_i - \boldsymbol{p}_j \|.
\end{align}

\subsection{Details of Algorithm and Pseudocode}
\label{app:pseudocode}

\paragraph{\bench} Algorithm~\ref{alg:bench} presents the details of our filtering process for \bench, where $l^-$ and $l^+$ represent the minimal and maximal frame intervals, respectively, to make the filtering more efficient. In this process, we define an angular range $\tau^-$ and select four levels (15, 30, 45, 60) to simulate progressively more challenging cases. The predefined maximal central point deviation $\bar{d}$ is set to 300, as most images have a width of 640, and 300 is less than half of that.

\begin{algorithm}[tb]
    \footnotesize
    \caption{Abstract generation of \bench.}
    \label{alg:bench}
    \begin{algorithmic}[1]
        \REQUIRE{\( \{(I_i, D_i, \mathbf{K}_i, \mathbf{T}_{wc_i})\}_{i= 1, \dots, n}, \tau^-, d^+, l^-, l^+ \).}
        \ENSURE{$I_i, I_j, \boldsymbol{v}^*, \tau^*, \bar{d}^*$}
        
        \FOR{\(i = 1, \dots, n-1\)}
            \FOR{\(j = i+l^-, \dots, \min(n, i+l^+)\)}
                \STATE Calculate $\tau$ and $\bar{d}$ in reprojection.
                \IF{\(\tau \in [\tau^-, \tau^-+5]\) and \(\bar{d} < d^+\)}
                    \STATE $\mathbf{T}_{c_ic_j} \gets (\mathbf{T}_{wc_i})^{-1}\mathbf{T}_{wc_j}$.
                    \STATE Decompose $\mathbf{R}, \mathbf{t}$ from $\mathbf{T}_{c_ic_j}$.
                    \STATE Solve $\boldsymbol{v}$ from $\mathbf{R}, \mathbf{t}$.
                    \STATE $\tau^* \gets \tau$.
                    \STATE $\bar{d}^* \gets \bar{d}$.
                \ENDIF
            \ENDFOR
        \ENDFOR
    \end{algorithmic}
\end{algorithm}

\paragraph{\diag} Algorithm~\ref{alg:diag} presents the details of our filtering process for \diag. To filter images that meet our requirements, we select meaningful thresholds that allow the images to show differences while still being part of the same scene. We use the following set of thresholds $\{(\delta_i^-, \delta_i^+)\}_{i = 1, \dots, 6}$, as shown in \autoref{tab:th}.

\begin{table}[tb]
    \centering
    \begin{tabular}{l|rr}
         & $\delta^-$ & $\delta^+$ \\
    \midrule
       $\theta$ & 5 & 15 \\
       $\phi$  & 5 & 15 \\
       $\psi$  & 3 & 10 \\
       $t_x$  & 0.15 & 0.4 \\
       $t_y$  & 0.1 & 0.3 \\
       $t_z$  & 0.15 & 0.4 \\
    \end{tabular}
    \caption{Thresholds used in \diag}
    \label{tab:th}
\end{table}

\begin{algorithm}[tb]
    \footnotesize
    \caption{Abstract generation of \diag.}
    \label{alg:diag}
    \begin{algorithmic}[1]
        \REQUIRE{\( \{(I_i, \mathbf{T}_{wc_i})\}_{i= 1, \dots, n}, \{(\delta_i^-, \delta_i^+)\}_{i = 1, \dots, 6}, l^-, l^+\).}
        \ENSURE{$I_i, I_j, \boldsymbol{v}^*$.}
        
        \FOR{\(i = 1, \dots, n-1\)}
            \FOR{\(j = i+l^-, \dots, \min(n, i+l^+)\)}
                \STATE $\mathbf{T}_{c_ic_j} \gets (\mathbf{T}_{wc_i})^{-1}\mathbf{T}_{wc_j}$.
                \STATE Decompose $\mathbf{R}, \mathbf{t}$ from $\mathbf{T}_{c_ic_j}$.
                \STATE Solve $\boldsymbol{v}$ from $\mathbf{R}, \mathbf{t}$.
                \FOR{\(k = 1, \dots, 6\)}
                    \IF{\(|\boldsymbol{v}_k| > \delta_k^+\)}
                        \IF{\(|\boldsymbol{v}_{m \neq k}| < \delta_m^-\)}
                            \STATE $\boldsymbol{v}^* \gets \boldsymbol{v}$.
                        \ENDIF
                    \ENDIF
                \ENDFOR
            \ENDFOR
        \ENDFOR
    \end{algorithmic}
\end{algorithm}

\subsection{Dataset Statistics}
\label{app:data-stat}

Table~\ref{tab:stat-diag} gives the statistics about \diag. Table~\ref{tab:bench-comp} compares the general spatial reasoning benchmark with ours, highlighting our challenging setup.

\begin{table}[tb]
    \centering
    \small
    \input{tab/stat_diag}
    \caption{Statistics for \diag.}
    \label{tab:stat-diag}
\end{table}

\begin{table}[tb]
    \centering
    \resizebox{\linewidth}{!}{
        \setlength{\tabcolsep}{2pt}
        \input{tab/bench_comparison}
    }
    \caption{Comparison of spatial reasoning benchmarks by task categories (RU - Relation Understanding, ME - Metric Estimation, CG - Coordinate Grounding, PP - Path Plan, SR - Scene Reasoning, Rot. - Rotation, Tr. - Translation, Comb. - Combination, Cont. - Continuous, Disc. - Discrete), highlighting the novelty and focus of our work on spatial reasoning research.
    }
    \label{tab:bench-comp}
\end{table}

\section{Experimental Setup}
\label{app:exp-setup}

\subsection{Baseline Evaluation}
\label{app:baseline-eval}

We use SIFT and LoFTR to detect matching points across images, then apply RANSAC to compute the relative pose, which is represented in homogeneous pose matrix format. From this, we extract the rotation matrix and translation vector. The relative pose vector $\boldsymbol{v}$ is then obtained, and we check the sign of the translation along the y-axis. This is because we aim to maintain the same label space as in VLM—binary classification. Therefore, we evaluate the performance of the computer vision methods based on the sign of the y-axis translation in their predicted values.

\subsection{VLM Evaluation}
\label{app:vlm-eval}

To enhance the reasoning ability of VLMs, we use a special output format instead of a greedy strategy. Specifically, we want the VLM to reason first within the \verb|<thinking>| tags and then provide the answer within the \verb|<ans>| tags. To minimize bias, we also randomly shuffle the order of answer candidates for each sample.

The example prompts we use for \bench and for \diag can be found in \autoref{fig:ex-prompt-bench} and \autoref{fig:ex-prompt-diag}, respectively.

\begin{figure*}[tb]
    \centering
\begin{tcolorbox}[
    title=Example prompt for VLM on \bench,
    boxrule=1pt,
    left=0mm, right=0mm, top=0mm, bottom=0mm,
    fontupper=\ttfamily,
]
\begin{lstlisting}[language=]
<input>
You are given two consecutive frames from a scene. The first image shows the **source viewpoint**, and the second image shows the **target viewpoint**. The camera moves and rotates simultaneously between these two frames. Use the visual difference between them to reason about the camera motion in 3D space.
</input>

<task>
Select the correct description of the camera motion between these viewpoints.
</task>

<ans-candidates>
0. Move left while yawing right
1. Move right while yawing left
</ans-candidates>

<output-format>
Please stricly follow the format: Provide your spatial reasoning inside <thinking></thinking> XML tags, and provide **only index of your option** inside <ans></ans> XML tags, e.g., <thinking>...</thinking> <ans>...</ans>.
Do not output anything else.
</output-format>
\end{lstlisting}
\end{tcolorbox}
    \caption{Example prompt for VLM on \bench.}
    \label{fig:ex-prompt-bench}
\end{figure*}

\begin{figure*}[tb]
    \centering
\begin{tcolorbox}[
    title=Example prompt for VLM on \diag,
    boxrule=1pt,
    left=0mm, right=0mm, top=0mm, bottom=0mm,
    fontupper=\ttfamily,
]
\begin{lstlisting}[language=]
<input>
You are given two consecutive frames from a scene. The first image shows the **source viewpoint**, and the second image shows the **target viewpoint**. The camera movement is usually described in 6 degree of freedom (6DoF)---translation and rotation along x, y, and z axes. We now have a constrain on camera motion between source and target viewpoints, that is, the significant movement is only rotation along y-axis (yaw), i.e., rotate left or right.
</input>

<task>
Use the visual difference between them to reason about the direction of this significant camera motion in 3D space. Select the correct description of the camera motion between these viewpoints.
</task>

<ans-candidates>
0. Rotate right
1. Rotate left
</ans-candidates>

<output-format>
Please stricly follow the format: Provide your spatial reasoning inside <thinking></thinking> XML tags, and provide **only index of your option** inside <ans></ans> XML tags, e.g., <thinking>...</thinking> <ans>...</ans>.
Do not output anything else.
</output-format>
\end{lstlisting}
\end{tcolorbox}
    \caption{Example prompt for VLM on \diag.}
    \label{fig:ex-prompt-diag}
\end{figure*}

\subsection{Human Evaluation}
\label{app:human-eval}

We randomly select samples from \bench and design a Google Form to collect data. The same prompt is used for human annotators, with the exception of special tokens and formatting, to minimize any potential bias.

The prompt is then presented to human annotators (see \autoref{fig:ex-prompt-human}), with the options provided in the Google Form.

\begin{figure*}[tb]
    \centering
\begin{tcolorbox}[
    title=Prompt for human annotators,
    boxrule=1pt,
    left=0mm, right=0mm, top=0mm, bottom=0mm,
    fontupper=\ttfamily,
]
\begin{lstlisting}[language=]
You are given two consecutive frames from a scene. The first image shows the source viewpoint, and the second image shows the target viewpoint. The camera moves and rotates simultaneously between these two frames. Use the visual difference between them to reason about the camera motion in 3D space.
\end{lstlisting}
\end{tcolorbox}
    \caption{Prompt for human annotators.}
    \label{fig:ex-prompt-human}
\end{figure*}

\section{Main Results}
\label{app:main-res}

\subsection{Case Study on \bench}
\label{app:case-study}

As shown in \autoref{fig:case}, \gptf accurately detects local objects, yet fails to describes their relative shifts and infer the correct camera motion between viewpoints rigorously. 
This confusion stems from the inherent ambiguity between camera rotation and translation, revealing a deeper misalignment between visual perception and 3D spatial reasoning. 

\begin{figure*}[tb]
    \centering
    \input{fig/case_study}
    \caption{Qualitative examples from \gptf on \bench. 
    Although the model correctly identifies objects, it fails to ground relative shifts and reason about the underlying 3D camera motion (right), revealing the inherent ambiguity between rotation and translation.}
    \label{fig:case}
\end{figure*}

\subsection{Full Results on \diag}
\label{app:tab-diag}

\autoref{tab:res-diag} shows the full numbers for \autoref{fig:res-diag}.

\begin{table*}[tb]
    \centering
    \resizebox{.65\linewidth}{!}{
        \input{tab/diag_res}
    }
    \caption{Full table of results on \diag. See \autoref{fig:res-diag} for visualization about top-performing VLMs.}
    \label{tab:res-diag}
\end{table*}

\section{Error Analysis}
\label{app:ea}

\subsection{Full Results and Visualization for Intra-Image Analysis}
\label{app:ea-i}

\begin{figure*}[tb]
    \centering
    \includegraphics[width=\linewidth]{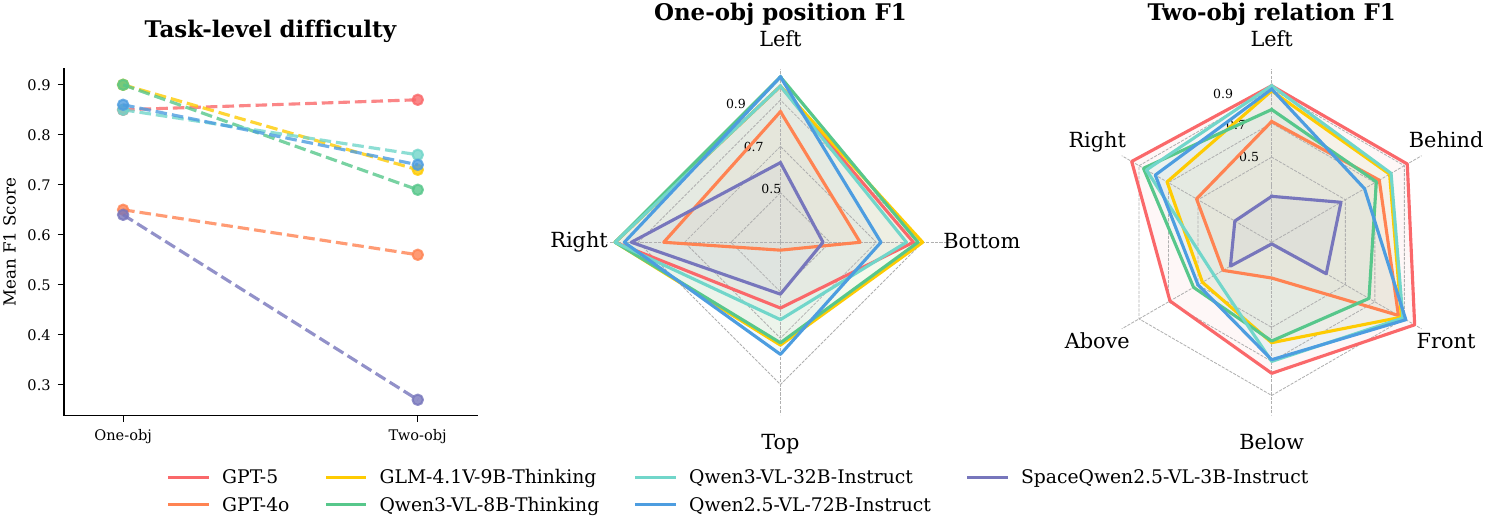}
    \caption{Evaluating intra-image spatial reasoning. This stage ensures that models possess the requisite spatial vocabulary before moving to multi-view geometric tasks.}
    \label{fig:ea-i}
\end{figure*}

We show the full table of VLMs on WhatsUp dataset in \autoref{tab:whatsup-long}. Also, \autoref{fig:ea-i} present the visualization about how VLMs perform on our annotated dataset, splited by each spatial relation.

\begin{table*}[tb]
    \centering
    \resizebox{\linewidth}{!}{
        \input{tab/whatsup_long}
    }
    \caption{Performance on the \texttt{WhatsUp} dataset (following their metrics). We compare current frontier VLMs against previous state-of-the-art models (e.g., BLIP-2) to establish a baseline for single-image spatial competence.}
    \label{tab:whatsup-long}
\end{table*}

\subsection{Full Results for Cross-Image Analysis}
\label{app:ea-ii}

\autoref{tab:ea-ii} shows the full numbers for \autoref{fig:ea-ii}.

\begin{table}[tb]
    \centering
    \setlength{\tabcolsep}{2pt}
    \resizebox{\linewidth}{!}{
        \input{tab/ea_ii}
    }
    \caption{Full table for cross-image task. See \autoref{fig:ea-ii} for visualization.}
    \label{tab:ea-ii}
\end{table}

%% file: tab/stat_diag.tex
\begin{tabular}{lrrrrrr}
\toprule
 & \multicolumn{6}{c}{DoF} \\
 \cmidrule(lr){2-7}
 & $\theta$ & $\phi$ & $\psi$ & $t_x$ & $t_y$ & $t_z$ \\
\midrule
\#~Frames & 285 & 123 & 258 & 143 & 29 & 53 \\
\bottomrule
\end{tabular}

%% file: tab/bench_comparison.tex
\begin{tabular}{l*{10}{c}}
\toprule
 & \multicolumn{3}{c}{\textbf{3D}} & \multicolumn{2}{c}{\textbf{Mental}} & \multicolumn{3}{c}{\textbf{Camera}} & \multicolumn{2}{c}{\textbf{Multi-view}} \\

\cmidrule(lr){2-4} \cmidrule(lr){5-6} \cmidrule(lr){7-9} \cmidrule(lr){10-11}

 & \textbf{RU} & \textbf{ME} & \textbf{CG} & \textbf{PP} & \textbf{SR} & \textbf{Rot.} & \textbf{Tr.} & \textbf{Comb.} & \textbf{Cont.} & \textbf{Disc.} \\ 

\midrule

{WhatsUp} & \gm & \rx & \rx & \rx & \rx & \rx & \rx & \rx & \rx & \rx \\

{BLINK} & \gm & \rx & \rx & \rx & \gm & \rx & \gm & \rx & \gm & \gm \\

{Spatial-MM} & \gm & \rx & \rx & \rx & \gm & \rx & \rx & \rx & \rx & \rx \\

{SpatialEval} & \rx & \rx & \rx & \rx & \gm & \rx & \rx & \rx & \rx & \rx \\

{Q-Spatial} & \rx & \gm & \rx & \rx & \rx & \rx & \rx & \rx & \rx & \rx \\

{SpatialRGPT} & \gm & \gm & \rx & \rx & \gm & \rx & \rx & \rx & \rx & \rx \\

{MM-Spatial} & \gm & \gm & \gm & \rx & \rx & \rx & \rx & \rx & \gm & \rx \\

{RoboSpatial} & \gm & \rx & \gm & \rx & \gm & \rx & \rx & \rx & \rx & \rx \\

{VSI-Bench} & \gm & \gm & \rx & \gm & \gm & \rx & \rx & \rx & \gm & \rx \\

{MindCube} & \rx & \rx & \rx & \rx & \gm & \rx & \rx & \rx & \rx & \gm \\

{LRR-Bench} & \gm & \rx & \rx & \rx & \rx & \gm & \gm & \rx & \gm & \rx \\

\midrule

\textbf{\bench} & \rx & \rx & \rx & \rx & \rx & \rx & \rx & \gm & \gm & \rx \\

\textbf{\diag} & \rx & \rx & \rx & \rx & \rx & \gm & \gm & \rx & \gm & \rx \\

\bottomrule
\end{tabular}

%% file: fig/case_study.tex
\begingroup
\footnotesize
\setstretch{0.9}

\definecolor{g1}{HTML}{99e2b4}
\definecolor{g2}{HTML}{88d4ab}
\definecolor{g3}{HTML}{78c6a3}
\definecolor{g4}{HTML}{67b99a}
\definecolor{g5}{HTML}{56ab91}
\definecolor{g6}{HTML}{469d89}
\definecolor{g7}{HTML}{358f80}

\definecolor{r1}{HTML}{ffccd5}
\definecolor{r2}{HTML}{ffb3c1}
\definecolor{r3}{HTML}{ff8fa3}
\definecolor{r4}{HTML}{ff758f}
\definecolor{r5}{HTML}{ff4d6d}
\definecolor{r6}{HTML}{c9184a}

\newtcolorbox{pairbox}[1][]{
  enhanced,
  colback=white,
  colframe=gray!50,
  boxrule=0pt,
  boxsep=0pt,
  left=0pt, right=0pt, top=0pt, bottom=0pt,
  sharp corners,
  title={#1}, 
  fonttitle=\bfseries\small,
  coltitle=black,
  colbacktitle=white,
  attach boxed title to top center,
  boxed title style={empty, top=0pt, bottom=0pt, boxrule=.8pt},
}

\newtcolorbox{ques}[3][]{
enhanced,title=Prompt,
left=0pt, right=0pt, top=0pt, bottom=0pt,
watermark graphics=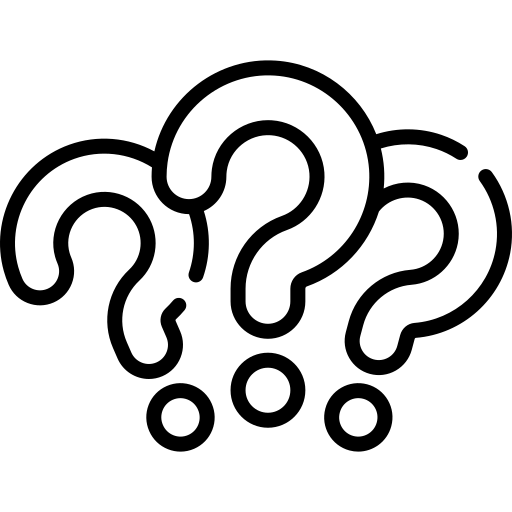,
watermark opacity=0.15,
arc=1mm,
colback=#2, 
colframe=#3, 
colbacktitle=#2, 
fonttitle=\bfseries,
coltitle=#3, 
attach boxed title to top left={yshift=-0.50mm,xshift=4pt},
boxed title style={
skin=enhancedfirst jigsaw,
size=fbox,
boxrule=.5mm,
arc=1mm,
bottom=-1mm,
interior style={
fill=none,
top color=#2,
bottom color=#2,
},
},
#1
}

\newtcolorbox{ans}[4][]{
enhanced,title=#4,
left=0pt, right=0pt, top=0pt, bottom=0pt,
watermark graphics=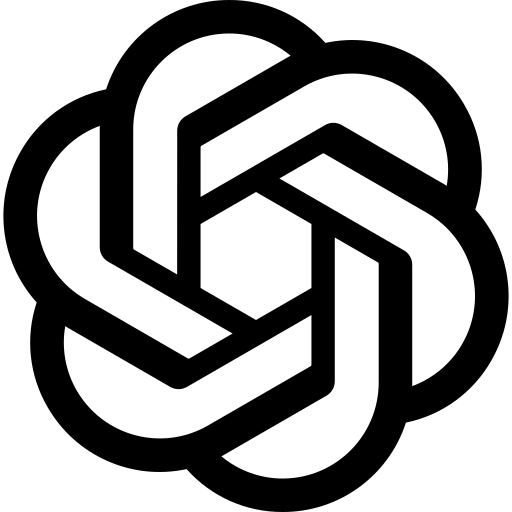,
watermark opacity=0.05,
arc=1mm,
colback=#2, 
colframe=#3, 
colbacktitle=#2, 
fonttitle=\ttfamily\bfseries,
coltitle=#3, 
attach boxed title to top left={yshift=-0.50mm,xshift=4pt},
boxed title style={
skin=enhancedfirst jigsaw,
size=fbox,
boxrule=.5mm,
arc=1mm,
bottom=-1mm,
interior style={
fill=none,
top color=#2,
bottom color=#2,
},
},
#1
}

\newtcolorbox{com}[3][]{
enhanced,title=Comments,
left=0pt, right=0pt, top=0pt, bottom=0pt,
arc=1mm,
colback=#2, 
colframe=#3, 
colbacktitle=#2, 
fonttitle=\bfseries,
coltitle=#3, 
attach boxed title to top left={yshift=-0.50mm,xshift=4pt},
boxed title style={
skin=enhancedfirst jigsaw,
size=fbox,
boxrule=.5mm,
arc=1mm,
bottom=-1mm,
interior style={
fill=none,
top color=#2,
bottom color=#2,
},
},
#1
}

\newtcbox{\mybox}[1][red]{
on line,
arc=0pt,outer arc=0pt,colback=#1!10!white,colframe=#1!50!black,
boxsep=0pt,
left=0pt, right=0pt, top=1pt, bottom=1pt,
boxrule=0pt,bottomrule=1pt,toprule=1pt}

\newtcbox{\xmybox}[1][red]{on line,
arc=7pt,colback=#1!10!white,colframe=#1!50!black,
before upper={\rule[-3pt]{0pt}{10pt}},boxrule=1pt,
boxsep=0pt,
left=0pt, right=0pt, top=1pt, bottom=1pt,
}

\begin{tcbraster}[
    raster columns=2,
    raster equal height,
]
\begin{tcolorbox}[
colback=g2!60,
boxrule=0pt,
left=2mm, right=2mm, top=0pt, bottom=0pt,
]
    \begin{pairbox}[Correct Case]
        \centering
        \begin{minipage}[t]{0.48\linewidth}
          \centering
          \includegraphics[width=\linewidth]{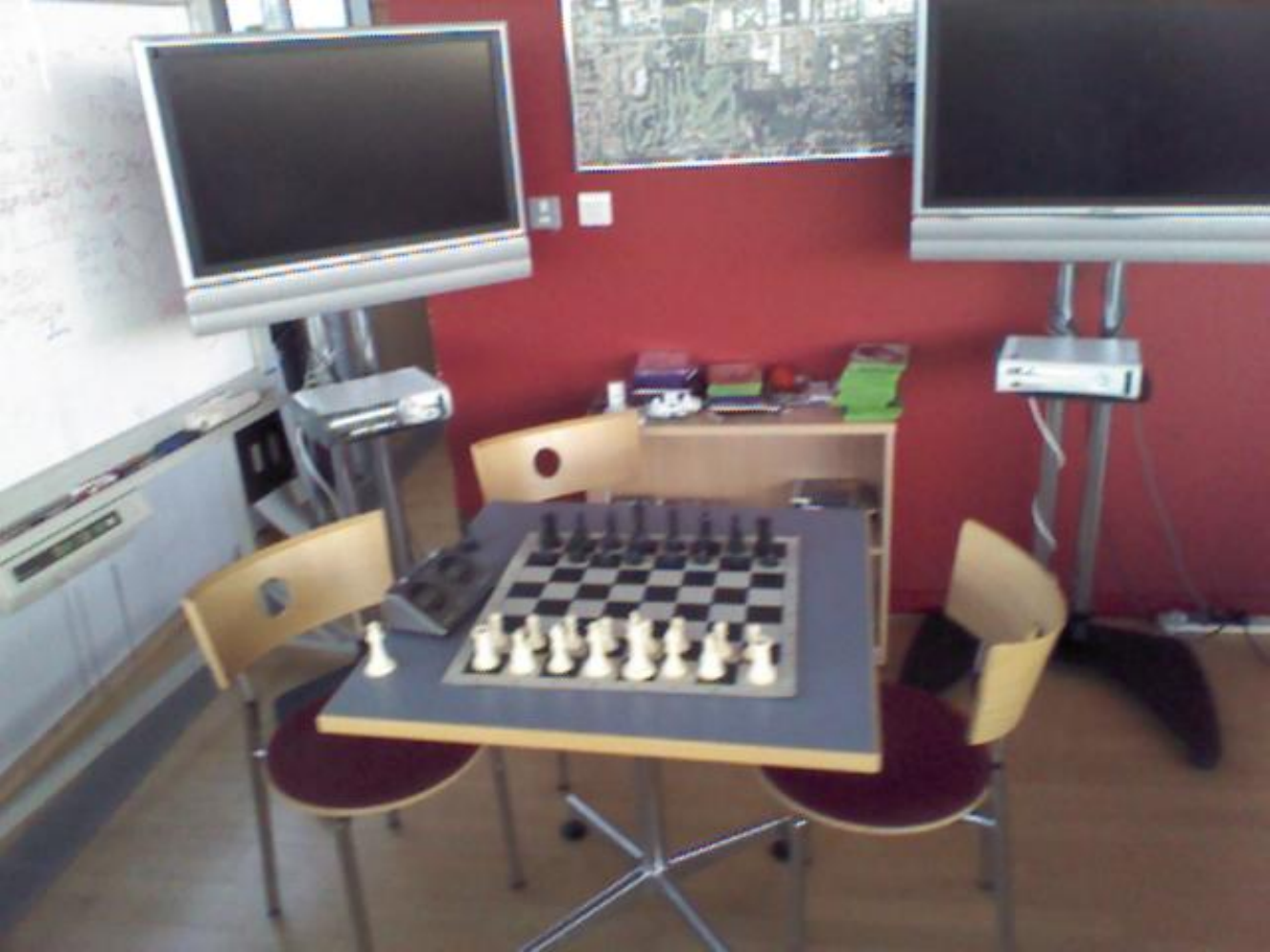}
        \end{minipage}
        \begin{minipage}[t]{0.48\linewidth}
          \centering
          \includegraphics[width=\linewidth]{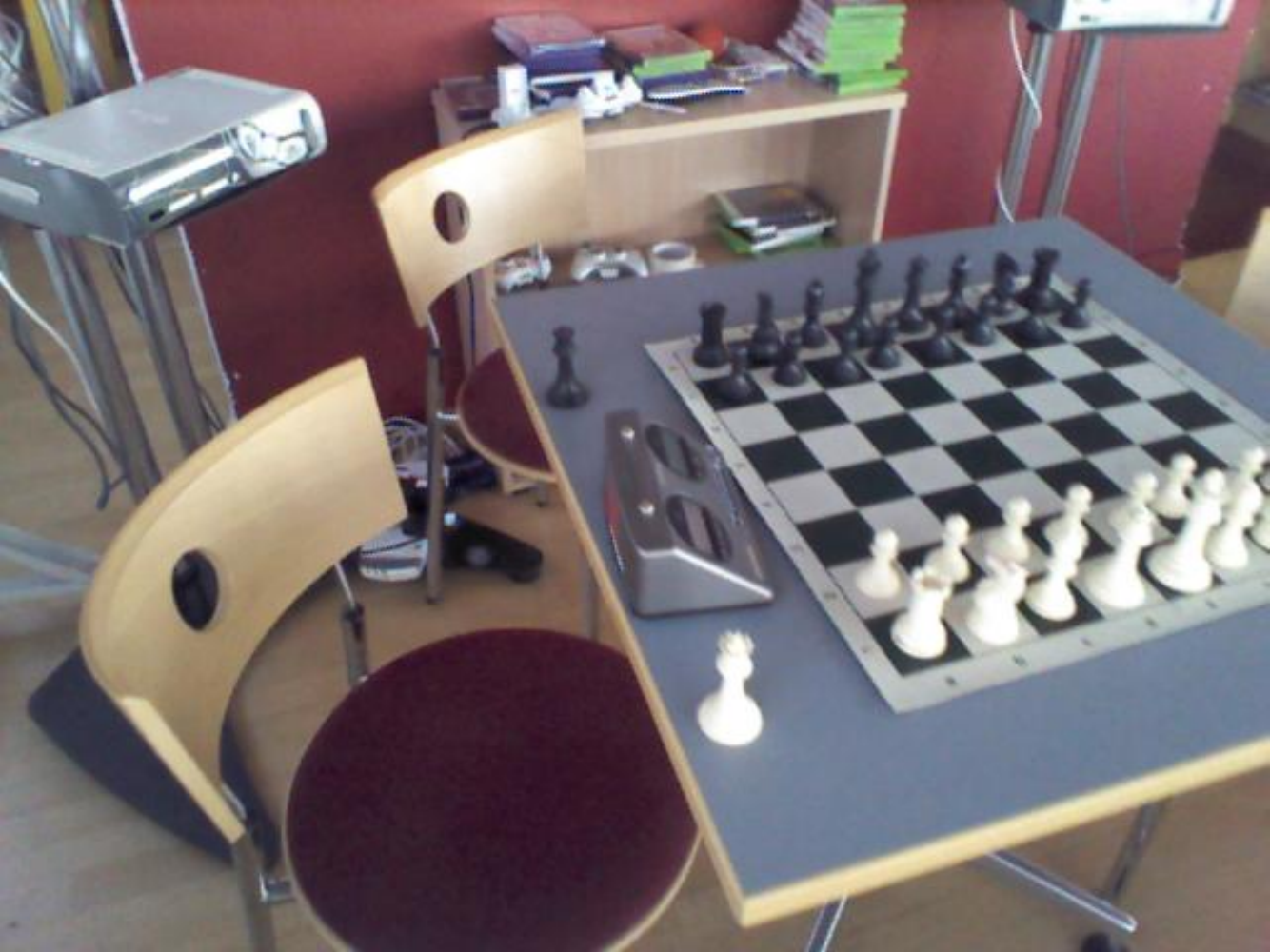}
        \end{minipage}
    \end{pairbox}
    \begin{ques}{g1!10}{g5!80!black}
    0. Move left while yawing right
    
    1. Move right while yawing left
    \end{ques}
    \begin{ans}{g1!10}{g5!80!black}{GPT-5}
    <thinking>
    
    Between frames, the nearby \mybox[DarkOrchid]{chess table and chairs} \mybox[Lime]{shift right} in the image, indicating the \mybox[DeepSkyBlue]{camera translated left} (near objects move opposite the motion). The \mybox[DarkOrchid]{chessboard’s perspective} shows depth receding \mybox[Lime]{toward the upper-right}, meaning the camera is now \mybox[DeepSkyBlue]{pointed more to the right} than before (yaw right). Hence: move left while yawing right.
    
    </thinking>
    
    <ans>0</ans>
    \end{ans}
    \begin{com}{g1!10}{g5!80!black}
        \mybox[DarkOrchid]{Excellent object detection};\mybox[Lime]{Excellent visual perception};
        
        \mybox[DeepSkyBlue]{Excellent reasoning on perspective-transformation}
    \end{com}
\end{tcolorbox}
\begin{tcolorbox}[
colback=red!20, boxrule=0pt,
left=2mm, right=2mm, top=0pt, bottom=0pt,
]
    \begin{pairbox}[Incorrect Case]
        \centering
        \begin{minipage}[t]{0.48\linewidth}
          \centering
          \includegraphics[width=\linewidth]{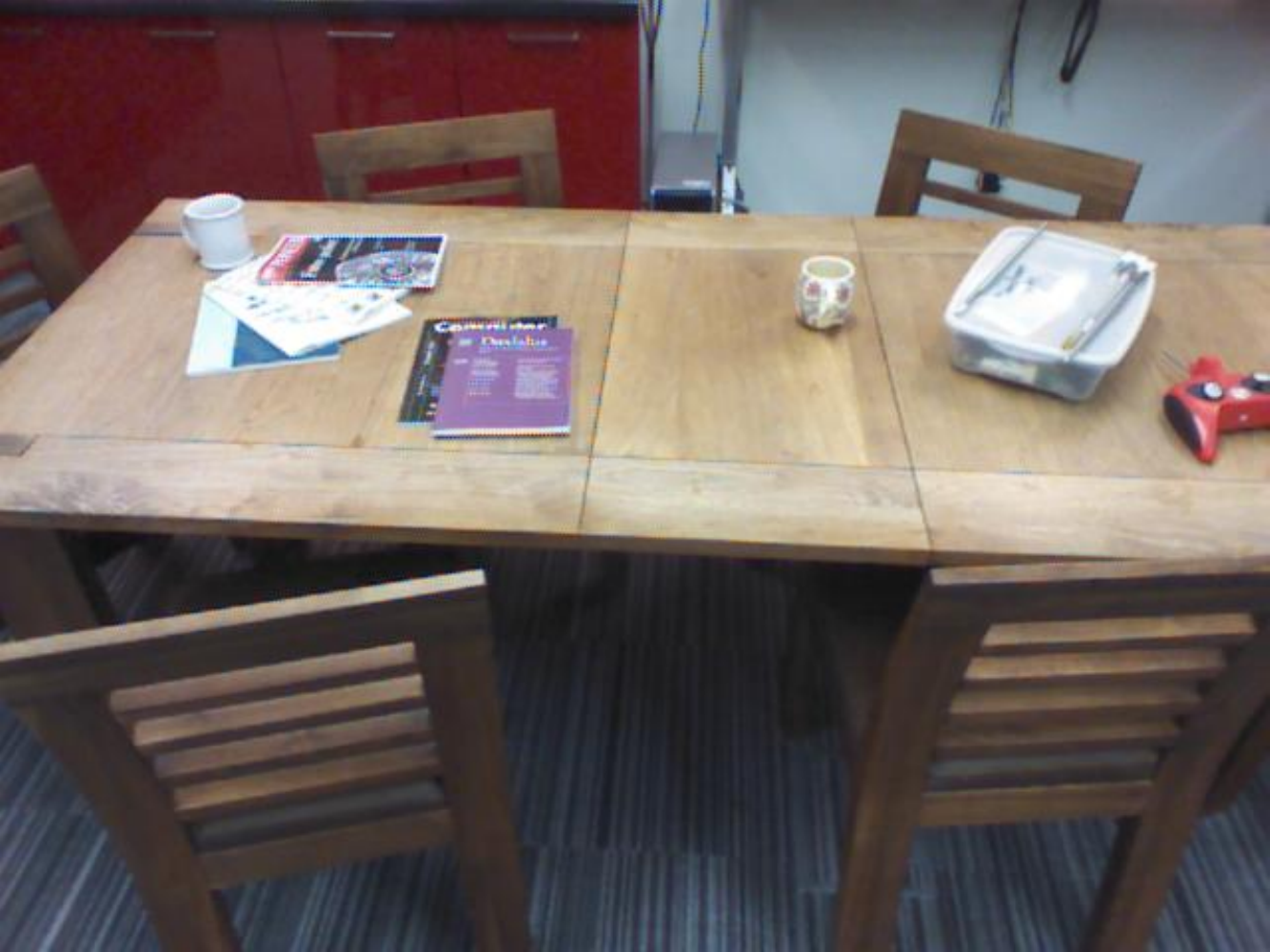}
        \end{minipage}
        \begin{minipage}[t]{0.48\linewidth}
          \centering
          \includegraphics[width=\linewidth]{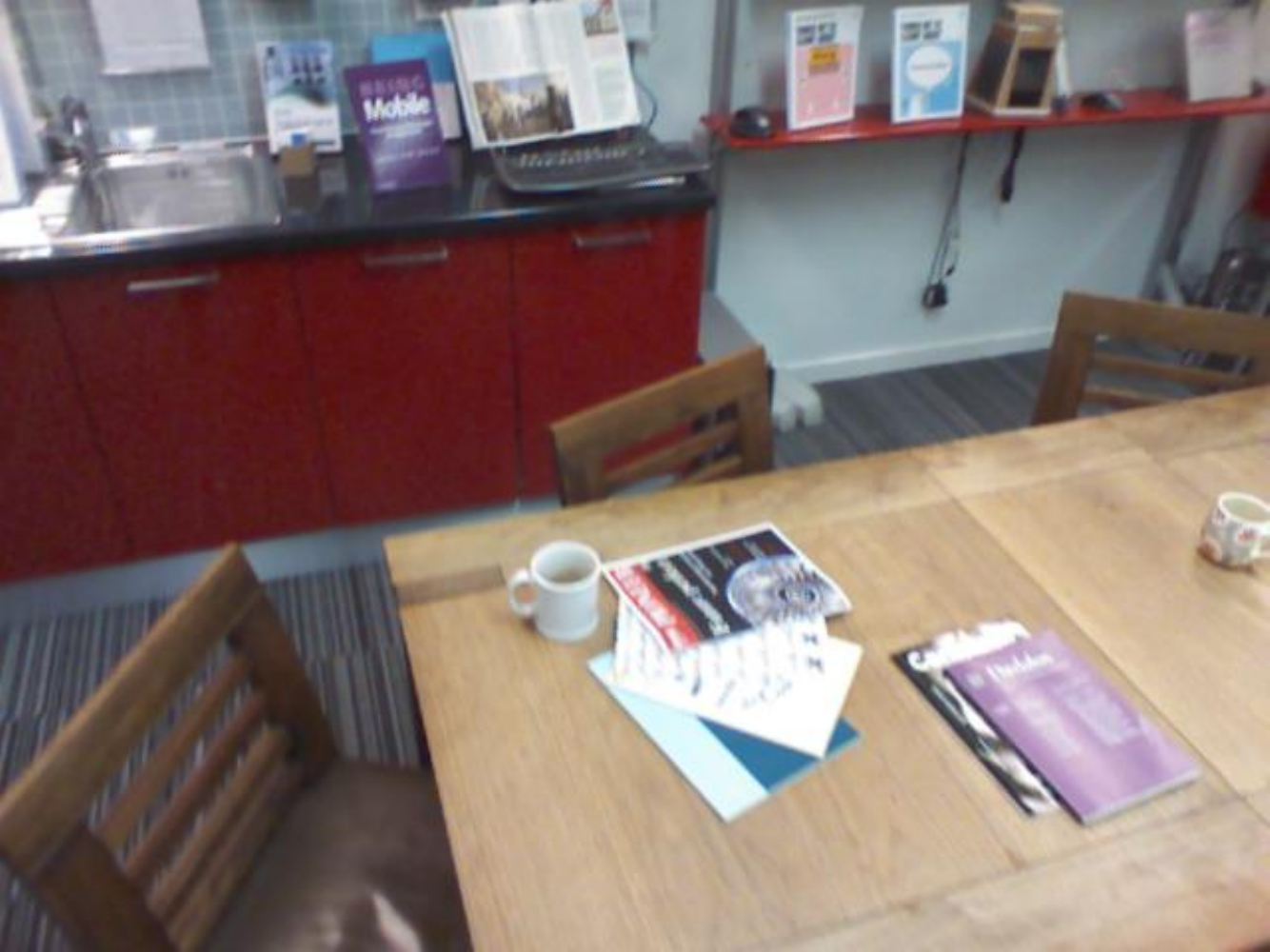}
        \end{minipage}
    \end{pairbox}
    \begin{ques}{r1!10}{r5!80!black}
    0. Move left while yawing right
    
    1. Move right while yawing left
    \end{ques}
    \begin{ans}{r1!10}{r5!80!black}{GPT-5}
    <thinking>
    
    In the target frame, more of the \mybox[DarkOrchid]{sink and left cabinets} \mybox[Lime]{come into view} while the \mybox[DarkOrchid]{right wall/shelf} \mybox[Lime]{moves toward the right} edge. This global shift to the right indicates \mybox[DeepSkyBlue]{a yaw to the left}. At the same time, nearby objects (\mybox[DarkOrchid]{table edge, chair}) \mybox[Lime]{shift left} relative to the distant background, showing opposite-direction parallax consistent with a \mybox[DeepSkyBlue]{rightward translation}. Thus the camera moved right while yawing left.
    
    </thinking>
    
    <ans>1</ans>
    \end{ans}
    \begin{com}{r1!10}{r5!80!black}
        \mybox[DarkOrchid]{Excellent object detection}; 
        \mybox[Lime]{Table and chairs appear to shift right};
        
        \mybox[DeepSkyBlue]{Inaccurate reasoning on perspective-transformation}
    \end{com}
\end{tcolorbox}
\end{tcbraster}
\endgroup

%% file: tab/diag_res.tex
\begin{tabular}{lrrrrrrr}
\toprule
 & $\theta$ & $\phi$ & $\psi$ & $t_x$ & $t_y$ & $t_z$ & Avg. \\
\midrule
llama3-llava-next-8b-hf & 0.59 & 0.55 & 0.52 & 0.61 & 0.38 & 0.77 & 0.58 \\
llava-onevision-qwen2-7b-ov-hf & 0.46 & 0.32 & 0.54 & 0.38 & 0.37 & 0.37 & 0.42 \\
Idefics3-8B-Llama3 & 0.42 & 0.58 & 0.48 & 0.54 & 0.67 & 0.57 & 0.52 \\
SpaceQwen2.5-VL-3B-Instruct & 0.54 & 0.48 & 0.55 & 0.54 & 0.41 & 0.64 & 0.52 \\
Qwen2.5-VL-3B-Instruct & 0.51 & 0.54 & 0.52 & 0.61 & 0.46 & 0.70 & 0.55 \\
Qwen2.5-VL-7B-Instruct & 0.59 & 0.60 & 0.49 & 0.60 & 0.73 & 0.67 & 0.59 \\
Qwen2.5-VL-32B-Instruct & 0.86 & 0.78 & 0.51 & 0.64 & 0.76 & 0.69 & 0.74 \\
Qwen2.5-VL-72B-Instruct & 0.86 & 0.78 & 0.53 & 0.62 & 0.69 & 0.73 & 0.74 \\
Qwen3-VL-4B-Instruct & 0.78 & 0.73 & 0.48 & 0.68 & 0.69 & 0.67 & 0.70 \\
Qwen3-VL-8B-Instruct & 0.68 & 0.84 & 0.49 & 0.75 & 0.63 & 0.74 & 0.72 \\
Qwen3-VL-32B-Instruct & 0.98 & 0.97 & 0.57 & 0.88 & 0.90 & 0.87 & 0.90 \\
Qwen3-VL-8B-Thinking & 0.79 & 0.80 & 0.49 & 0.62 & 0.68 & 0.56 & 0.71 \\
GLM-4.1V-9B-Thinking & 0.87 & 0.90 & 0.51 & 0.78 & 0.83 & 0.74 & 0.81 \\
GPT-4o & 0.97 & 0.77 & 0.37 & 0.72 & 0.86 & 0.64 & 0.77 \\
GPT-5 & 0.99 & 0.99 & 0.47 & 0.91 & 0.83 & 0.94 & 0.90 \\
\bottomrule
\end{tabular}

%% file: tab/whatsup_long.tex
\begin{tabular}{l*{11}{c}}
\toprule
 & \multicolumn{3}{c}{\textbf{What'sUp Subset A}} & \multicolumn{3}{c}{\textbf{What'sUp Subset B}} & \multicolumn{2}{c}{\textbf{COCO-spatial}} & \multicolumn{2}{c}{\textbf{GQA-spatial}} &  \\
 & \textbf{Indiv.} & \textbf{Pairs} & \textbf{Set of 4} & \textbf{Indiv.} & \textbf{Pairs} & \textbf{Set of 4} & \textbf{One-obj} & \textbf{Two-obj} & \textbf{One-obj} & \textbf{Two-obj} & \textbf{Avg.} \\
\midrule
BLIP2-ITM & 44.9 & 24.3 & 0.0 & 30.4 & 2.0 & 0.0 & 48.3 & 57.7 & 46.0 & 53.6 & 46.8 \\
BLIP2-ITC & 35.9 & 3.4 & 0.0 & 22.1 & 0.0 & 0.0 & 55.6 & 51.8 & 52.6 & 49.5 & 44.6 \\
\midrule

Idefics3-8B-Llama3 & 53.9 & 36.4 & 2.9 & 51.0 & 27.9 & 5.9 & 75.9 & 67.0 & 78.8 & 69.4 & 66.0 \\ 

llama3-llava-next-8b-hf & 25.5 & 4.3 & 0.0 & 22.3 & 3.9 & 0.0 & 48.6 & 53.0 & 48.8 & 57.4 & 42.6 \\

llava-onevision-qwen2-7b-ov-hf & 91.0 & 84.5 & 70.9 & 74.0 & 56.9 & 25.5 & 85.0 & 82.5 & 87.5 & 87.3 & 84.5 \\

SpaceQwen2.5-VL-3B-Instruct & 83.3 & 69.4 & 44.7 & 71.6 & 23.1 & 1.9 & 68.2 & 65.7 & 68.2 & 67.4 & 70.7 \\

Qwen2.5-VL-3B-Instruct & 89.3 & 82.0 & 66.0 & 73.8 & 60.3 & 27.5 & 92.2 & 87.7 & 93.0 & 88.7 & 87.4 \\

Qwen2.5-VL-7B-Instruct & 86.2 & 76.7 & 53.4 & 83.6 & 71.1 & 49.0 & 94.0 & 89.8 & 94.3 & 83.2 & 88.5 \\ 

Qwen2.5-VL-32B-Instruct & 98.1 & 96.6 & 93.2 & 87.7 & 79.9 & 59.8 & 93.3 & 91.8 & 94.7 & 89.0 & 92.5 \\ 

Qwen2.5-VL-72B-Instruct & 99.0 & 98.5 & 97.1 & 95.6 & 91.7 & 83.3 & 93.9 & 96.1 & 96.0 & 95.2 & 96.2 \\ 

Qwen3-VL-4B-Instruct & 99.5 & 99.5 & 99.0 & 97.8 & 95.6 & 91.2 & 89.0 & 82.0 & 90.6 & 87.3 & 91.0 \\ 

Qwen3-VL-8B-Instruct & 96.6 & 93.7 & 87.4 & 95.3 & 91.2 & 82.4 & 93.6 & 83.9 & 95.1 & 89.0 & 92.3 \\ 

Qwen3-VL-32B-Instruct & 99.5 & 99.5 & 99.0 & 99.3 & 98.5 & 97.1 & 89.6 & 84.8 & 94.4 & 91.8 & 93.2 \\ 

GLM-4.1V-9B-Thinking & 99.0 & 98.5 & 97.1 & 96.8 & 94.1 & 88.2 & 96.2 & 95.9 & 97.2 & 98.6 & 97.3 \\

GPT-4o & 82.8 & 70.9 & 48.5 & 62.7 & 35.3 & 5.9 & 62.8 & 63.9 & 65.9 & 56.4 & 65.7 \\ 

GPT-5 & 99.5 & 99.5 & 99.0 & 99.5 & 99.0 & 98.0 & 96.8 & 96.8 & 97.3 & 98.6 & 98.1 \\ 
\midrule
Random chance & 25.0 & 6.3 & 0.4 & 25.0 & 6.3 & 0.4 & 50.0 & 50.0 & 50.0 & 50.0 & 41.7 \\
\bottomrule
\end{tabular}

%% file: tab/ea_ii.tex
\begin{tabular}{lrr}
\toprule
 & w/o Ref. & w/ Ref. \\
\midrule
llama3-llava-next-8b-hf & 0.31 & 0.30 \\
llava-onevision-qwen2-7b-ov-hf & 0.26 & 0.28 \\
Idefics3-8B-Llama3 & 0.24 & 0.30 \\
SpaceQwen2.5-VL-3B-Instruct & 0.26 & 0.26 \\
Qwen2.5-VL-3B-Instruct & 0.29 & 0.31 \\
Qwen2.5-VL-7B-Instruct & 0.28 & 0.29 \\
Qwen2.5-VL-32B-Instruct & 0.30 & 0.30 \\
Qwen2.5-VL-72B-Instruct & 0.30 & 0.31 \\
Qwen3-VL-4B-Instruct & 0.19 & 0.25 \\
Qwen3-VL-8B-Instruct & 0.25 & 0.34 \\
Qwen3-VL-32B-Instruct & 0.25 & 0.28 \\
Qwen3-VL-8B-Thinking & 0.39 & 0.47 \\
GLM-4.1V-9B-Thinking & 0.34 & 0.46 \\
GPT-4o & 0.19 & 0.24 \\
GPT-5 & 0.58 & 0.56 \\
\bottomrule
\end{tabular}